%% file: main.tex
\documentclass[10pt,a4paper]{article}

\usepackage[
  a4paper,
  top=2.5cm,
  bottom=3cm,
  left=2.5cm,
  right=2.5cm
]{geometry}
\usepackage{braket}
\usepackage[T1]{fontenc}
\usepackage{newtxtext,newtxmath}
\usepackage{amsmath,mathtools}
\usepackage[dvipsnames]{xcolor}
\usepackage{graphicx}
\usepackage{booktabs}
\usepackage{subcaption}
\usepackage{microtype}
\usepackage{titlesec}
\usepackage{enumitem}
\usepackage{fancyhdr}
\usepackage[most]{tcolorbox}
\usepackage{tikz}
\usepackage[numbers,sort&compress]{natbib}
\usepackage{hyperref}
\usepackage{array}
\usepackage{tabularx}
\usepackage[normalem]{ulem}
\usetikzlibrary{arrows.meta,positioning,shapes.geometric,fit,calc}

\definecolor{THUBaseLight}{HTML}{A69BBD}
\definecolor{THUBaseDark}{HTML}{3A216D}
\colorlet{THULightTint}{THUBaseLight!10}
\colorlet{THUDarkTint}{THUBaseDark!10}
\colorlet{THUAccent}{THUBaseDark}
\definecolor{THUGray}{HTML}{5D6675}
\definecolor{THUDark}{HTML}{1C2330}

\colorlet{ReportAccent}{THUAccent}
\colorlet{ReportDark}{THUDark}
\colorlet{ReportGray}{THUGray}
\colorlet{ReportTint}{THUDarkTint}
\colorlet{ReportBlueTint}{THULightTint}
\definecolor{ReportGreenTint}{HTML}{F3F8F4}

\newcommand{\thunlpArticleType}{}

\newcommand{\setthunlparticlemeta}[1]{%
  \renewcommand{\thunlpArticleType}{#1}%
}
\newcommand{\thunlpauthoraffildivider}{%
  {\color{THUAccent}\rule{0.35\linewidth}{1.5pt}}%
}
\newcommand{\thunlpAffiliation}{}
\newcommand{\thunlpaffiliation}[1]{%
  \gdef\thunlpAffiliation{#1}%
}
\newcommand{\thunlpTitleImage}{}
\newcommand{\setthunlptitleimage}[1]{%
  \gdef\thunlpTitleImage{#1}%
}

\hypersetup{
  colorlinks=true,
  linkcolor=THUAccent,
  citecolor=THUAccent,
  urlcolor=THUAccent
}

\titleformat{\section}
  {\large\bfseries\color{THUAccent}}
  {\thesection}
  {0.5em}
  {}
\titleformat{\subsection}
  {\normalsize\bfseries\color{THUDark}}
  {\thesubsection}
  {0.5em}
  {}
\titlespacing*{\section}{0pt}{0.8em}{0.4em}
\titlespacing*{\subsection}{0pt}{0.55em}{0.3em}

\newif\ifthunlpLogoVisible
\thunlpLogoVisibletrue
\newcommand{\thunlplogooff}{\thunlpLogoVisiblefalse}

\newcommand{\thunlpHeaderLogo}{}

\fancypagestyle{plain}{%
  \fancyhf{}
  
  \fancyhead[L]{\thunlpHeaderLogo}
  \fancyhead[R]{}
  \fancyfoot[C]{\textcolor{THUGray}{\thepage}}
}

\makeatletter
\def\@maketitle{%
  \vspace*{-0.7em}
  \noindent
  \begin{minipage}[c]{0.75\textwidth}
    \raggedright
    {\setlength{\fboxsep}{5pt}%
     \colorbox{THUDarkTint}{%
       \textbf{\small\color{THUAccent}\thunlpArticleType}%
     }}\par
    \vspace{0.55em}
    {\fontsize{20}{24}\selectfont
     \bfseries\color{THUDark}\@title\par}
  \end{minipage}%
  \ifx\thunlpTitleImage\@empty\else
    \hfill
    \begin{minipage}[c]{0.23\textwidth}
      \centering
      \includegraphics[height=70pt]{\thunlpTitleImage}
    \end{minipage}%
  \fi
  \par\vspace{0.5em}
  \noindent{\normalsize\@author\par}
  \ifx\thunlpAffiliation\@empty\else
    \par\vspace{0.24em}
    \noindent{\normalsize\parbox[t]{\textwidth}{\thunlpAffiliation}\par}
  \fi
  \vspace{0.16em}
  {\small\color{THUGray}\@date\par}
  \par\vspace{0.2em}
  \@thanks
  \thispagestyle{fancy}
}
\newcommand{\makefrontmatter}[1]{%
  \if@twocolumn
    \twocolumn[
    \maketitle
    \begin{abstract}
    #1
    \end{abstract}
    \vspace{0.25em}
    ]
  \else
    \maketitle
    \begin{abstract}
    #1
    \end{abstract}
    \vspace{0.25em}
  \fi
}
\makeatother

\renewenvironment{abstract}{%
  \begin{tcolorbox}[
    enhanced,
    colback=THULightTint,
    colframe=THUAccent,
    boxrule=0.8pt,
    arc=2pt,
    left=7pt,right=7pt,top=6pt,bottom=6pt,
    before skip=0.25em,
    after skip=0.8em
  ]
  {\bfseries\color{THUAccent}Abstract}\hspace{0.55em}
}{%
  \end{tcolorbox}
}

\setlist[itemize]{leftmargin=1.8em,itemsep=0.18em,topsep=0.25em}
\setlist[enumerate]{leftmargin=2.0em,itemsep=0.22em,topsep=0.25em}
\renewcommand{\arraystretch}{1.12}
\usepackage{multirow}
\usepackage{colortbl}
\usepackage{siunitx}
\usepackage{wrapfig}

\title{Learning Individual Scientific Taste \\ with Privileged Hindsight Distillation}
\date{}
\setthunlparticlemeta{RESEARCH ARTICLE}
\thunlplogooff

\setthunlptitleimage{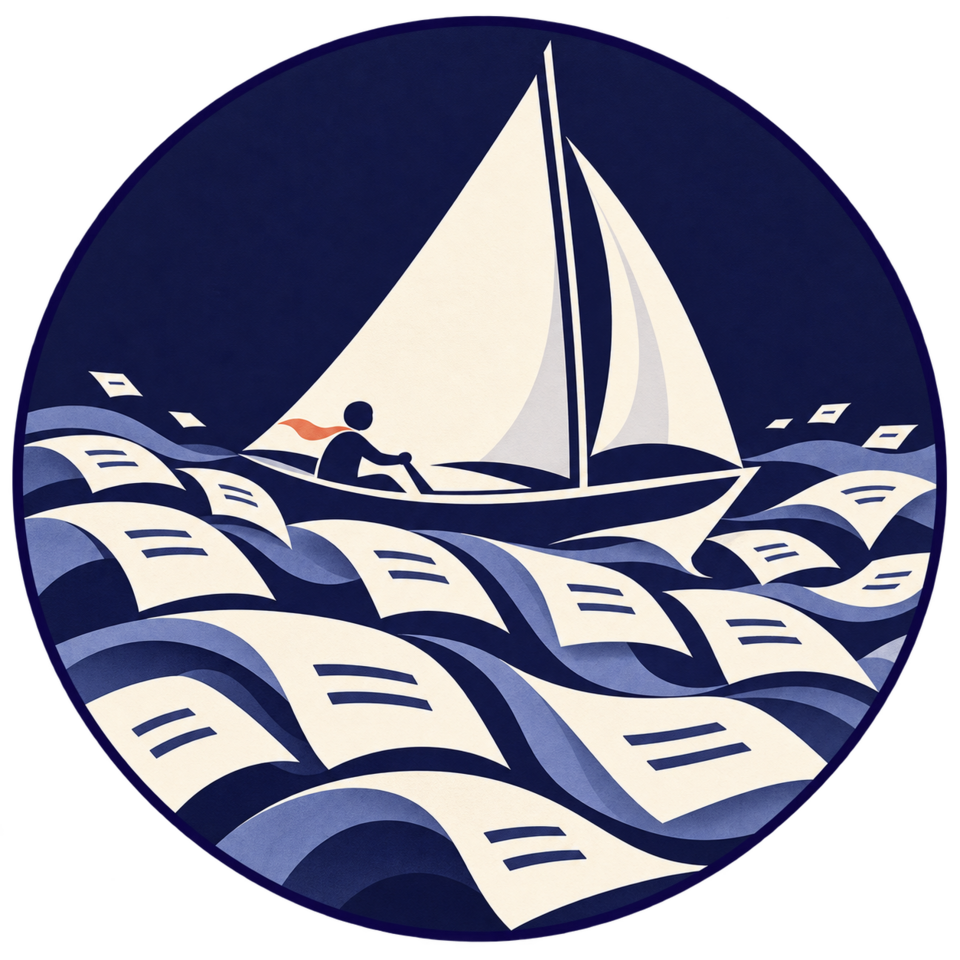}

\author{
  \textbf{Hanwei Fan\textsuperscript{1,2}}\quad
  \textbf{Yanqiao Wang\textsuperscript{1,3,4}}\quad
  \textbf{Yiqi Zhu\textsuperscript{1,2}}\quad
  \textbf{Xuanyu Lei\textsuperscript{1,2}}\quad
  \textbf{Peng Li\textsuperscript{1,$*$}}\quad
  \textbf{Yang Liu\textsuperscript{1,2,$*$}}\quad \\
  \thunlpauthoraffildivider
}
\thunlpaffiliation{
  \textsuperscript{1}Institute for AI Industry Research (AIR), Tsinghua University, Beijing, China\\  
  \textsuperscript{2}Department of Computer Science and Technology, Tsinghua University, Beijing, China\\
  \textsuperscript{3}Yau Mathematical Sciences Center, Tsinghua University, Beijing, China \\
  \textsuperscript{4}Qiuzhen College, Tsinghua University, Beijing, China \\
}

\begin{document}
\makefrontmatter{
Scientific taste shapes which research directions researchers pursue, and it varies across individuals. Existing work evaluates research directions along multiple evaluation criteria or models scientific taste using community feedback signals such as citations and reviews, but neither directly captures how individuals translate assessments under the same criteria into different preferences. We model scientific taste as a researcher-specific mapping from criterion-level assessments in a shared, evidence-grounded evaluation space with proposal-visible evidence, to relative preferences. A population-level mapping learned from retrospective supervision provides a cold start when individual feedback is scarce. We introduce Privileged Hindsight Distillation (PHD), which uses later-emerging scientific evidence during training to improve proposal-time criterion assessment and the population mapping, while exposing only proposal-visible information at inference. We evaluate our approach primarily in quantum research, where structured public reviews are largely unavailable, and use pairwise proposal comparison as a direct interface for observing individual taste. With only 15 comparisons per researcher, our method captures sharply different tastes, including opposing preferences over the same research-direction pairs when they trade off across evaluation criteria. In an interactive study, personalized rankings also surface proposals associated with greater subsequent user engagement, connecting learned taste to downstream research exploration.
}

\begingroup
\renewcommand{\thefootnote}{\fnsymbol{footnote}}
\footnotetext[1]{Corresponding authors: \href{mailto:lipeng@air.tsinghua.edu.cn}{lipeng@air.tsinghua.edu.cn}, \href{mailto:liuyang2011@tsinghua.edu.cn}{liuyang2011@tsinghua.edu.cn}.}
\endgroup

\input{introduction}
\section{Related Work}
\label{sec:related}

\paragraph{Modeling Scientific Taste.}
Prior work measures research taste through citation-based impact \citep{koltun2021taste}, or learns scientific judgment from citation comparisons \citep{tong2026taste}, institutional publication decisions \citep{gong2026institutional}, and future-impact supervision for peer review \citep{rasool2026reviewguard}. More recent work has moved scientific judgment toward earlier research stages: TASTE \citep{baig2026taste} benchmarks expert pairwise preferences of AI-Safety research proposals, while other studies compare human and LLM-generated ideas using a research-taste taxonomy \citep{chen2026gap} or explicitly encode researcher-specific preferences \citep{zhang2026kkanbu}. These studies examine aggregate judgment, proposal-level preferences, or differences in ideation patterns, whereas our focus is learning researcher-specific preferences from sparse feedback.

\paragraph{Research Artifact Evaluation.} The rapid development of scientific ideation systems \citep{wang2024scimon,baek2025researchagent,si2025can,lu2024aiscientist,gottweis2025coscientist} has made the evaluation of ideas, proposals, and papers increasingly important \citep{ding2026agentsurvey,ye2025designspace}. Existing benchmarks assess research artifacts along criteria such as novelty, feasibility, and potential impact \citep{wang2026ligbench,jiang2026hindsight,zhao2026apres,qiao2026innoeval}, yet holistic LLM judgments do not always predict the quality of executed research \citep{si2025ideationexecution} and show particular limitations in novelty assessment \citep{li2026novetylimits}. Evidence-grounded methods such as OpenNovelty \citep{zhang2026opennovelty} instead support assessment with retrieved literature and explicit evidence. Our work builds on this direction by grounding proposal-time criterion assessment in proposal claims and relevant literature, and organizing these assessments within a shared evaluation space.

\paragraph{Privileged Information Learning.}
Learning using privileged information formalizes supervision that is available during training but absent at deployment \citep{vapnik2009lupi,lopez2015unifying}. Recent language-model methods similarly distill privileged information into deployment-restricted policies \citep{yang2026selfdistilled,zhao2026opsd}; OPSD in particular lets a privileged teacher guide an unprivileged student on its own rollouts. Reflection and textual optimization methods improve agentic artifacts like prompts and LM programs \citep{yuksekgonul2024textgrad,agrawal2026gepa,hu2024mipro}, reusable experience \citep{zhao2024expel}, and episodic memory \citep{shinn2023reflexion}. Our proposed PHD connects these two categories by exposing retrospective scientific evidence only to the teacher while distilling its benefit into a proposal-time agentic workflow.

\paragraph{Sparse Preference Learning.}
Personalized preference learning studies how user-specific judgments can be recovered from sparse supervision, including few-shot adaptation and population-informed preference learning \citep{poddar2024vpl,singh2025fspo,li2026coldstart}. Relevant work models heterogeneous preferences through structured preference representations \citep{shenfeld2025rewardfactorization,bose2025lore}, while perspectivist learning preserves annotator-specific disagreement rather than collapsing judgments into a single consensus \citep{davani2021disagreements,xu2026perspectivist,orlikowski2025beyond,sorensen2024pluralistic}. When feedback arrives sequentially, continual-learning methods commonly use replay to incorporate new observations while mitigating forgetting \citep{chaudhry2019tiny,aljundi2019mir}. Our setting brings these two problems together by learning individual scientific taste from sparse, sequential researcher feedback.

\input{method}

\input{experiments}

\section{Conclusion}

We propose a framework for learning individual scientific taste from sparse researcher feedback, representing research directions in a shared, evidence-grounded evaluation space and adapting a population-level preference mapping to individual researchers. Privileged Hindsight Distillation uses later-emerging evidence during training to improve proposal-time assessment and population-level learning without exposing hindsight at inference. Experiments reveal substantial researcher-specific variation, including opposing preferences over the same proposal pairs, and show that conditional rules capture preference patterns beyond fixed criterion reweighting in real researchers. In interactive deployment, personalized rankings also surface proposals associated with greater subsequent engagement than the matched Single Call baseline. While our study is limited by sparse real-expert feedback and a finite set of evaluation criteria, the results suggest that population-level judgment provides a useful starting point but does not fully capture individual scientific preferences.

\bibliography{refs}
\bibliographystyle{unsrtnat}

\appendix{}
\input{appendix}

\end{document}

%% file: introduction.tex
\section{Introduction}

Scientific taste broadly refers to the judgment that shapes what researchers choose to work on, how they approach scientific problems, and how they recognize the potential impact of research ideas and findings~\citep{mitchison2013taste,tjian2023judgment,tong2026taste}. As large language models (LLMs) increasingly support scientific research from hypothesis and proposal generation to experimental design, execution, and paper drafting~\citep{lu2024aiscientist,yamada2025aiscientistv2,schmidgall2025agentlab,gottweis2025coscientist}, generating research directions becomes more scalable, making it increasingly important to decide which are worth pursuing before their outcomes are known.

Scientific taste, however, varies across researchers. Researchers differ in which problems they consider important and which approaches they prefer~\citep{mitchison2013taste}, and substantial disagreement persists even when domain experts evaluate the same scientific work~\citep{bornmann2010reliability,mutz2012heterogeneity}. Learning individual taste is particularly challenging when direct feedback is scarce. Recent work evaluates research directions along criteria such as novelty, significance, and feasibility~\citep{qiao2026innoeval,pei2026deepinstructor}, but criterion-level assessments alone do not specify how a particular researcher translates them into preferences. Other work learns scientific judgment at scale from ecosystem signals such as citations, public reviews, and publication outcomes~\citep{koltun2021taste,tong2026taste,gong2026institutional,rasool2026reviewguard}. When aggregated, these signals provide population-level supervision rather than the taste of any particular individual.

\begin{figure*}[t]
\centering
\includegraphics[width=\textwidth]{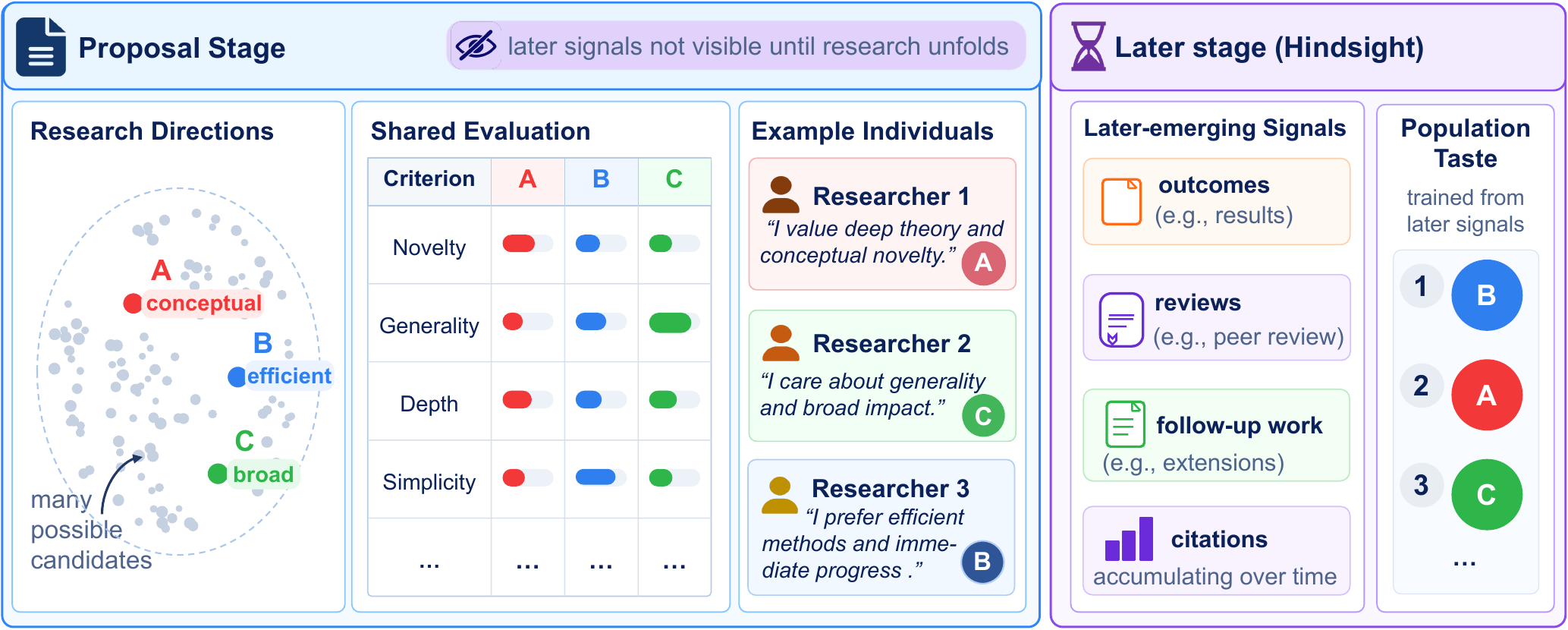}
\caption{\textbf{Modeling scientific taste.} Proposal-stage evidence places research directions in a shared evaluation space, while researchers may map the same criterion profiles to different preferences. Later-emerging evidence can provide retrospective supervision for population-level learning but is unavailable at proposal-time inference.}
\label{fig:teaser}
\end{figure*}

To model this variation, we separate shared evaluation from individual preference. At the proposal stage, each research direction is represented in a shared evaluation space over criteria such as novelty and generality, with assessments grounded in proposal claims and relevant literature~\citep{zhang2026opennovelty}. This common representation supports comparison across directions. Population-level learning can be instantiated in this shared space to map criterion-level assessments to aggregate preferences, while individual taste is modeled as a researcher-specific mapping from criterion-level assessments and proposal-visible evidence to relative preferences (Figure~\ref{fig:teaser}). The same shared evaluation can therefore support different choices: one researcher may favor conceptual novelty, while another prioritizes broad applicability. Although these criteria are neither fully objective nor exhaustive of scientific judgment, they provide a common basis for modeling variation across researchers.

A separate challenge is that most useful supervision signals emerge after research unfolds. Outcomes, community reviews, follow-up work and citations can reveal how a direction is received, but such later-emerging signals are unavailable when directions must be judged at the proposal stage. Their availability is also uneven across fields: computer science venues such as ICLR expose structured public reviews through OpenReview~\citep{soergel2013openscholarship}, while comparable records are more limited in quantum research, where platforms such as SciRate~\citep{scirate} provide weaker community signals (scites). When aggregated, these later signals can still support learning a population-level taste mapping, which provides a useful cold start for personalization under sparse direct feedback. We therefore use this mapping as a population-level prior for personalization, not assuming that it represents a unique community-wide taste.

To make such retrospective supervision usable at proposal time, we introduce \textbf{Privileged Hindsight Distillation (PHD)}. In PHD, a privileged teacher uses later-emerging evidence to refine the rubrics, conditional rules, and memory of a proposal-only student, improving both shared evaluation and population-level taste learning while preserving proposal-only inference. The resulting population mapping is then adapted to individual researchers using sparse direct feedback.

We observe individual scientific taste through pairwise proposal comparison, a lightweight interface for eliciting relative preferences without requiring absolute scores~\citep{dras2015pairwise,narimanzadeh2023pairwise,baig2026taste}. We evaluate our approach primarily in quantum research, with additional cross-domain experiments using real and simulated researchers. Our results show that PHD improves proposal-time evaluation and population-level prior learning, while sparse feedback enables adaptation to individual researchers. We further observe substantial disagreement across researchers, particularly when competing directions trade off across evaluation criteria, and deploy the complete framework in an interactive proposal exploration system, where personalized rankings more effectively surface proposals that users subsequently choose to inspect, refine, or pursue.

\paragraph{Contributions.}
\textbf{(1)} We formulate scientific taste by separating criterion-level assessment from how those assessments are valued. A shared, evidence-grounded evaluation space supports population and researcher-specific mappings to relative preferences, with the former serving as a transferable prior for individual adaptation.
\textbf{(2)} We introduce Privileged Hindsight Distillation (PHD), which uses later-emerging scientific evidence as privileged supervision to learn criterion-level assessment and the population prior while preserving proposal-only inference.
\textbf{(3)} We show that individual scientific taste can diverge substantially from population-level judgment. Researchers exhibit sharply different preferences on the same research-direction pairs, and researcher-specific conditional mappings recover conflicting preferences that fixed criterion reweighting misses under sparse feedback.

%% file: method.tex
\section{Method}
\label{sec:method}

\begin{figure*}[t]
\centering
\includegraphics[width=0.95\textwidth]{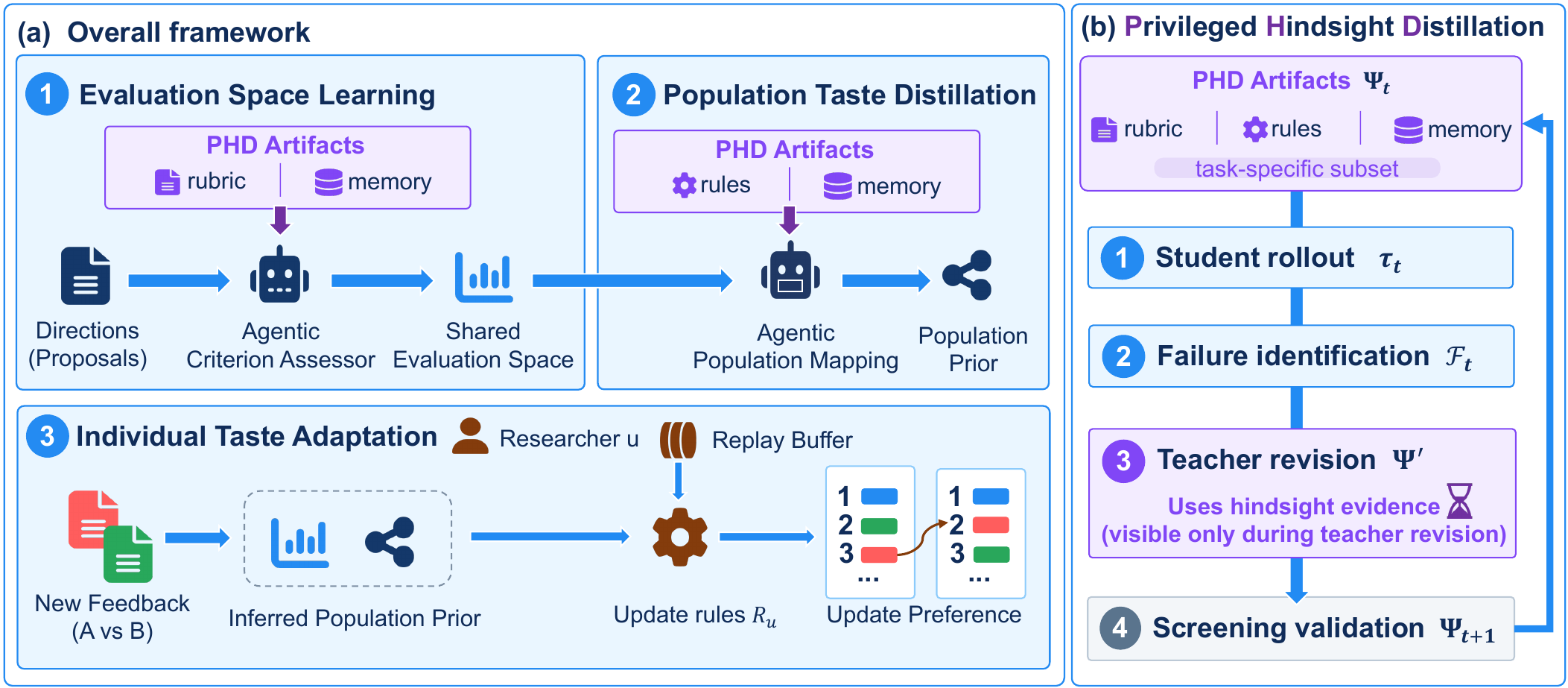}
\caption{
\textbf{Overview of our framework.}
(a) The framework learns a shared evaluation space, distills a population-level taste mapping, and sequentially adapts it to individual researchers; PHD refines the task-specific artifacts used by the first two components.
(b) PHD iteratively identifies student failures, uses privileged hindsight evidence to diagnose them and revise the artifacts, and retains only revisions that pass leakage screening and held-out validation. At inference, the student uses only proposal-visible information and the final accepted artifacts.
}
\label{fig:pipeline}
\end{figure*}

Our framework learns individual scientific taste from sparse feedback through three components, as illustrated in Figure~\ref{fig:pipeline}(a). \textbf{Evaluation Space Learning} maps each research direction to a researcher-independent representation consisting of criterion-level assessments and associated proposal-visible evidence. \textbf{Population Taste Distillation} learns a population-level mapping from this shared representation to relative preferences. \textbf{Individual Taste Adaptation} then adapts the population mapping into a researcher-specific taste mapping using sparse sequential feedback. The first two components are improved through Privileged Hindsight Distillation (PHD), a shared training procedure described in Section~\ref{sec:phd}.

\subsection{Shared Evaluation Space and Scientific Taste}
\label{sec:def}
Let \(\mathcal C\) be a finite set of candidate research directions, with \(P\in\mathcal C\) representing one direction through its proposal. All proposal-stage information, including the proposal and literature available at that time, is denoted by \(\mathcal I_{\mathrm{vis}}(P)\), with \(\mathcal E(P)\) representing the corresponding proposal-visible evidence. Both depend only on \(P\), not on the researcher evaluating it. We define a shared \(K\)-dimensional evaluation space \(\mathcal S\), with one dimension per evaluation criterion. A criterion assessor maps proposal-visible information to \[\hat{s}(P)=f_{\mathrm{crit}}(\mathcal I_{\mathrm{vis}}(P))=(\hat{s}_1(P),\ldots,\hat{s}_K(P))\in\mathcal S,\] where each \(\hat{s}_k(P)\) is researcher-independent. Researchers therefore share the same criterion representation but may translate it and its supporting evidence into different preferences.

For researcher \(u\), scientific taste is a researcher-specific mapping from criterion assessments and proposal-visible evidence to relative preferences,
\[
T_u:\{(\hat{s}(P),\mathcal E(P))\}_{P\in\mathcal C}\longmapsto \succ_u,
\]
where \(A\succ_u B\) means that \(u\) prefers \(A\) to \(B\). We observe scientific taste through pairwise comparisons. For \(A,B\in\mathcal C\), let \(x_{A,B}=\hat{s}(A)-\hat{s}(B)\) denote their criterion-difference vector and \(\mathcal E_{A,B}=(\mathcal E(A),\mathcal E(B))\) their proposal-visible evidence. We instantiate \(T_u\) through a scalar preference function \(U_u\), such that \[A\succ_u B\Longleftrightarrow U_u(x_{A,B},\mathcal E_{A,B})>0.\] Section~\ref{sec:population-taste} defines the shared functional form of \(U\), and Section~\ref{sec:individual-taste} personalizes it from researcher feedback.

\subsection{Privileged Hindsight Distillation}
\label{sec:phd}

Scientific judgments must be made at the proposal stage, before completed methods, experimental results, subsequent literature, citations, and community feedback become available. Once research unfolds, these later observations provide additional evidence for diagnosing errors in proposal-time judgments. We denote such later-emerging information by \(\mathcal I_{\mathrm{priv}}\) and introduce \textbf{Privileged Hindsight Distillation (PHD)} to exploit it during training while preserving proposal-only inference.

Inspired by On-Policy Self-Distillation (OPSD)~\citep{zhao2026opsd}, we adapt its iterative self-improvement principle to an agentic workflow. With the base model frozen, PHD transfers hindsight-derived knowledge by refining the external artifacts that govern the student instead of updating model parameters. We denote the artifact state at refinement round \(t\) by \(\Psi_t\), comprising a task-specific subset of rubric, conditional rules, and memory. Rubrics specify criteria, scoring anchors, and evidence requirements; conditional rules encode context-dependent adjustments; and memory stores reusable proposal-visible guidance.

As illustrated in Figure~\ref{fig:pipeline}(b), PHD provides a shared four-stage refinement procedure for Evaluation Space Learning and Population Taste Distillation. For either task, let \(z\) denote a training input and \(y^*(z)\) its training-only reference target; their specific forms are defined in the following sections.

\textbf{1. Student rollout.}
Given proposal-visible information and the current artifacts \(\Psi_t\), the student produces a rollout \(\tau_t(z)\) containing the task prediction \(\hat y_t(z)\), confidence, and proposal-visible supporting evidence. Neither \(y^*(z)\) nor \(\mathcal I_{\mathrm{priv}}\) is available to the student.

\textbf{2. Failure identification.}
The analyzer compares \(\hat y_t(z)\) with the reference target \(y^*(z)\) using the task-specific error criterion and collects failed cases into \(\mathcal F_t\). Each failed case contains \(z\), its rollout \(\tau_t(z)\), and \(y^*(z)\). At this stage, the analyzer only identifies prediction failures; it does not access the privileged hindsight evidence used to construct \(y^*(z)\) or diagnose their causes.

\textbf{3. Teacher revision.}
The training-only teacher receives \(\Psi_t\), \(\mathcal F_t\), and \(\mathcal I_{\mathrm{priv}}\).  At this stage, privileged hindsight is used to diagnose the identified failures and propose a candidate revision \(\Psi'_t\). For example, hindsight may reveal that a high novelty score was driven mainly by unsupported novelty claims in the proposal. The teacher may then tighten the corresponding evidence requirement in the rubric, introduce a conditional rule for such cases, or retain the resulting proposal-visible guidance in memory.

\textbf{4. Screening and validation.}
The candidate revision first undergoes leakage screening, which rejects artifacts containing privileged outcomes, reference answers, or conditions unavailable at the proposal stage. Revisions that pass screening are evaluated on held-out data using the task-specific validation metric and are accepted as \(\Psi_{t+1}\) when they satisfy the corresponding acceptance criterion.

Accepted artifacts are reused in the next refinement round. At inference, only the final accepted artifacts and proposal-visible information remain; reference targets, the analyzer, teacher, and privileged hindsight are removed. Sections~\ref{sec:merit-space} and~\ref{sec:population-taste} specify the task input, reference target, editable artifacts, failure criterion, and validation metric for the two shared components.

\subsection{Evaluation Space Learning}
\label{sec:merit-space}

Evaluation Space Learning instantiates the criterion assessor \(f_{\mathrm{crit}}\) from Section~\ref{sec:def}. We use a shared \(K=6\)-dimensional space covering generality, novelty, importance, parsimony, depth, and algorithmic complexity. Following rubric-based evaluation~\citep{kim2024prometheus} and evidence-grounded scientific assessment~\citep{zhang2026opennovelty}, we construct two rubrics over the same criteria and score scales but with different information access.

\textbf{Rubric A} is the fixed reference rubric. It uses both proposal-visible information \(\mathcal I_{\mathrm{vis}}(P)\) and privileged hindsight \(\mathcal I_{\mathrm{priv}}(P)\) to produce the reference criterion vector \(s^*(P)=(s_1^*(P),\ldots,s_K^*(P))\). \textbf{Rubric B} is the student rubric and a component of \(\Psi\). It uses only \(\mathcal I_{\mathrm{vis}}(P)\) and governs the criterion assessor producing \(\hat s(P)\). The details of the two rubrics are provided in Appendices~\ref{app:rubric-a} and~\ref{app:rubric-b}.

For PHD, the task instance is \(z=P\), the reference target is \(y^*(P)=s^*(P)\), and the rollout \(\tau_t(P)\) contains \(\hat s_t(P)\), confidence, and proposal-visible supporting evidence. The analyzer computes the per-proposal mean absolute error (MAE) between \(\hat s_t(P)\) and \(s^*(P)\) across the \(K\) criteria and uses it to identify high-error proposals for the failure set \(\mathcal F_t\). The editable artifacts are Rubric B and case memory. When the assessor repeatedly mis-scores a criterion by at least one score level, we revise the corresponding scoring anchor in Rubric B. Isolated failures are instead stored as proposal-visible case lessons in memory, each with a triggering condition and an absolute-level check. Held-out MAE determines whether a revision is retained. Complete prompts and an illustrative revision are provided in Appendices~\ref{app:merit-workflow-prompts} and~\ref{app:merit-case-study}.

\subsection{Population Taste Distillation}
\label{sec:population-taste}

For this PHD instantiation, the task input is a proposal pair \(z=(A,B)\). The student observes the criterion-difference vector \(x_{A,B}\) and proposal-visible evidence \(\mathcal E_{A,B}\) defined in Section~\ref{sec:def}, and predicts \(\hat y_t(A,B)\in\{A,B\}\). The corresponding population reference target is \(y^*(A,B)\), whose construction is detailed in Appendix~\ref{app:PT}.

We model population and individual preferences using a common preference function. Each conditional rule \(r\) is evaluated independently on a research direction using only its proposal-visible representation. Let \(a_r(P)\) denote its activation on \(P\); for a pair \((A,B)\), its contribution is \(\phi_r(x_{A,B},\mathcal E_{A,B})=a_r(A)-a_r(B)\). The preference function is
\[
U(x,\mathcal E;w,R)
=
w^\top x
+
\sum_{r\in R}
\tilde{\alpha}_r\,\phi_r(x,\mathcal E),
\]
where \(w\) captures a fixed weighting over criterion differences, \(R\) contains conditional rules, and \(\tilde{\alpha}_r\) is the fitted rule coefficient adjusted by held-out reliability. The linear term captures preferences expressible by fixed criterion reweighting, while the rules capture context-dependent judgments beyond it. This functional form is shared by the population and individual mappings.

We first fit \(w_{\mathrm{pop}}\) on population training pairs using \(y^*(A,B)\) as supervision and no conditional rules, yielding \(U_{\mathrm{pop}}^{(0)}(x)=w_{\mathrm{pop}}^\top x\). We then freeze \(w_{\mathrm{pop}}\) and apply PHD to learn context-dependent residual judgments. The editable artifacts are population rules \(R_{\mathrm{pop}}\) and memory: rules encode reusable context-dependent adjustments, while memory stores proposal-visible guidance for applying them. These artifacts jointly govern the context-dependent component represented by \(R_{\mathrm{pop}}\) in the preference function. The resulting population mapping is \[U_{\mathrm{pop}}(x,\mathcal E)=U(x,\mathcal E;w_{\mathrm{pop}},R_{\mathrm{pop}}).\]

StrictPair, defined in Section~\ref{sec:exp-individual}, identifies failed pairs for \(\mathcal F_t\) and serves as the held-out validation metric. The resulting \(U_{\mathrm{pop}}\) provides the population prior for individual adaptation. Complete workflow and implementation details are provided in Appendix~\ref{app:population-workflow-prompts}.

\subsection{Individual Taste Adaptation}
\label{sec:individual-taste}

Individual Taste Adaptation learns directly from sparse pairwise feedback from researcher \(u\), stored in a time-ordered replay buffer \(\mathcal B_u\). Each observation contains a proposal pair \((A,B)\), its representation \(x_{A,B}\) and \(\mathcal E_{A,B}\) defined in Section~\ref{sec:def}, and the researcher preference \(y_u(A,B)\in\{A,B\}\). Using the common preference function from Section~\ref{sec:population-taste}, we personalize the population mapping through an individual residual \(\Delta w_u\) and researcher-specific rules \(R_u\):
\[
U_u(x,\mathcal E)=U\!\left(x,\mathcal E;w_u,R_{\mathrm{pop}}\cup R_u\right),
\] where $w_u=w_{\mathrm{pop}}+\Delta w_u$.
Adaptation starts from $U_{\mathrm{pop}}$ with \(\Delta w_u=0\) and \(R_u=\emptyset\).

As feedback is appended to \(\mathcal B_u\), \(\Delta w_u\) is refitted using the frozen population score as an offset, allowing criterion weights to deviate from the population prior. Recurring leave-one-out errors that remain after this linear adaptation motivate researcher-specific conditional rules. Each candidate rule is evaluated through cross-fitted replay over \(\mathcal B_u\): in each leave-one-out fold, \(\Delta w_u\) and the rule coefficient are refitted on the remaining observations and evaluated on the held-out support comparison. A rule is retained only if replay yields a net predictive improvement without breaking previously correct preferences. Held-out query labels and privileged hindsight are never used during individual adaptation. Further fitting and rule-induction details are provided in Appendix~\ref{app:individual-workflow-prompts}.

%% file: experiments.tex
\section{Experiments}

\subsection{Learning Individual Scientific Taste}
\label{sec:exp-individual}

\paragraph{Data.}
Quantum serves as our primary evaluation domain. We reconstruct research directions as proposal-stage inputs containing the research problem, methodology, evaluation plan, expected contributions, and risks, while excluding later-emerging information. All compared methods receive only proposal-visible inputs; as with other pretrained LLM evaluations, we do not assume control over parametric knowledge acquired during pretraining.

Individual-taste supervision must be collected directly from researchers, and is therefore naturally scarce. Each of seven quantum-algorithm researchers provides 15 sequential support comparisons for adaptation and 10 held-out query pairs for evaluation, with a binary preference and confidence annotation for each comparison. We complement this real-researcher cohort with 100 simulated researchers: 50 derived from publication histories of prominent quantum researchers (Sim.-Q) and 50 generated through controlled variation in criterion trade-offs (Sim.-Ctrl.). Further construction and supervision details are provided in Appendix~\ref{app:corpus-construction}.

\paragraph{Evaluation metric.}
For pairwise preference prediction, we use \textbf{StrictPair (SP)}, which requires both presentation orientations of a proposal pair to be predicted correctly. For the \(i\)-th pair \((A_i,B_i)\),
\[
\mathrm{SP}_i
=
\mathbb{I}\!\left[
\hat y(A_i,B_i)=y^*(A_i,B_i)
\;\land\;
\hat y(B_i,A_i)=y^*(B_i,A_i)
\right],
\]
where \(\hat y\) and \(y^*\) denote the predicted and target preferences, respectively. For individual taste, we report \textbf{Regret-Weighted StrictPair (RW-SP)} to account for researcher confidence. Following regret-based pairwise weighting~\citep{halpern2025pairwise}, comparison \(i\) receives weight \(d_i=2|p_i-0.5|\), where \(p_i\) is the preference probability derived from the researcher's confidence annotation, detailed in Appendix~\ref{app:rwsp}. We compute \(\mathrm{RW\text{-}SP}=(\sum_i d_i\,\mathrm{SP}_i)/(\sum_i d_i)\).

\paragraph{Baselines.}
We compare personalization approaches with increasingly explicit use of researcher feedback. Unless otherwise specified, all methods use \texttt{gpt-5.5} as the backbone, and no privileged hindsight is available during individual adaptation or inference. \textbf{Single Call (SC)} directly places all support comparisons in the prompt when predicting each held-out preference, while \textbf{Profile Prompting (PP)} first summarizes them into a researcher profile. SynthesizeMe~\citep{ryan2025synthesizeme}, FERMI~\citep{kim2025fermi}, and CIPHER~\citep{gao2024cipher} represent agentic approaches that maintain or refine user-specific descriptions or preferences from interaction history. VPL~\citep{poddar2024vpl}, LoRe~\citep{bose2025lore}, and FSPO~\citep{singh2025fspo} instead learn parameterized user-specific preference or reward representations. Detailed configurations are in Appendix~\ref{app:baseline-config}.

\paragraph{Results.}
Table~\ref{tab:individual-main} shows that representing the same sparse feedback through in-context or agentic personalization remains less effective than explicitly learning preference structure. Our method performs best on both real and simulated researchers while keeping the backbone frozen, indicating that explicit preference structure can be introduced through a transferable population mapping and lightweight researcher-specific state without per-user model-parameter updates.

\begin{table*}[t]
\centering
\caption{\textbf{Individual scientific taste prediction.} RW-SP across real and simulated researchers.}
\label{tab:individual-main}
\small
\vspace{0.5em}

\setlength{\tabcolsep}{6.0pt}
\renewcommand{\arraystretch}{1.10}

\begin{tabular*}{\textwidth}{
@{\extracolsep{\fill}}
l
cc
cccc
ccc
}
\toprule
&
\multicolumn{2}{c}{\textbf{In-context}}
&
\multicolumn{4}{c}{\textbf{Agentic}}
&
\multicolumn{3}{c}{\textbf{Parametric}}
\\
\cmidrule(lr){2-3}
\cmidrule(lr){4-7}
\cmidrule(lr){8-10}

&
SC
&
PP
&
SynthesizeMe
&
FERMI
&
CIPHER
&
\textbf{Ours}
&
VPL
&
LoRe
&
FSPO
\\

\midrule

\textbf{Real} $\uparrow$
&
0.479
&
0.481
&
0.477
&
0.419
&
0.524
&
\textbf{0.693}
&
0.629
&
0.595
&
\underline{0.642}
\\

\textbf{Sim.} $\uparrow$
&
0.423
&
0.519
&
0.473
&
0.366
&
0.399
&
\textbf{0.591}
&
0.556
&
0.547
&
\underline{0.559}
\\

\bottomrule
\end{tabular*}
\end{table*}

\begin{table*}[t]
\centering
\caption{
\textbf{Robustness across backbones.}
E-MAE/E-OA denote MAE/Ordering Accuracy for Evaluation Space Learning, and P-SP denotes StrictPair for Population Taste Distillation. Shared-component entries report \textbf{Ours / w/o PHD}. For individual adaptation, we report RW-SP for our complete method and alternative baselines under the same backbone.
}
\label{tab:backbone-sensitivity}

\vspace{0.6em}

\small
\renewcommand{\arraystretch}{1.15}
\setlength{\tabcolsep}{5.5pt}

\begin{tabular*}{\textwidth}{
@{\extracolsep{\fill}}
lcccccc
}
\toprule
& \multicolumn{2}{c}{Evaluation Space}
& \multicolumn{1}{c}{Population}
& \multicolumn{3}{c}{Individual RW-SP $\uparrow$}
\\
\cmidrule(lr){2-3}
\cmidrule(lr){4-4}
\cmidrule(lr){5-7}

Backbone
& E-MAE $\downarrow$
& E-OA $\uparrow$
& P-SP $\uparrow$
& Ours
& PP
& CIPHER
\\
\midrule

\texttt{gpt-5.5}
& \textbf{0.183} / 0.216
& \textbf{0.721} / 0.713
& \textbf{0.833} / 0.667
& \textbf{0.693}
& 0.481
& 0.524
\\

\texttt{gpt-5.4-mini}
& \textbf{0.233} / 0.292
& \textbf{0.665} / 0.604
& \textbf{0.667} / 0.567
& \textbf{0.613}
& 0.407
& 0.332
\\

\texttt{claude-opus-5}
& \textbf{0.217} / 0.275
& \textbf{0.675} / 0.612
& \textbf{0.500} / 0.367
& \textbf{0.664}
& 0.634
& 0.516
\\

\texttt{gemini-3.6-flash}
& \textbf{0.216} / 0.258
& \textbf{0.699} / 0.662
& \textbf{0.767} / 0.367
& \textbf{0.647}
& 0.422
& 0.405
\\

\bottomrule
\end{tabular*}
\end{table*}

\subsection{Robustness and Sensitivity Analysis}
\label{sec:exp-robustness}

\paragraph{Shared-component data.}
We further evaluate whether PHD successfully distills privileged hindsight in the two shared components that support individual taste learning. Both Evaluation Space Learning and Population Taste Distillation use carefully curated, paper-disjoint splits. Constructing these evaluations requires controlled proposal-time reconstruction and privileged reference construction, while population comparisons additionally require temporally comparable and informative proposal pairs. A proposal may participate in multiple comparisons within the same split, but no underlying paper is shared across splits.

For Evaluation Space Learning, reference criterion vectors \(s^*(P)\) are constructed with the fixed Rubric A using both proposal-visible information and privileged hindsight evidence. For Population Taste Distillation, reference preferences \(y^*(A,B)\) are aggregated from a multi-family LLM-as-a-Judge panel given the same information. Since these are LLM-generated references rather than independent human labels, we use them to evaluate whether PHD can distill hindsight supervision into proposal-time judgments and for corresponding ablations, rather than for external benchmarking. Construction details are provided in Appendices~\ref{app:corpus-construction}, \ref{app:rubric-a}, and \ref{app:PT}.

\paragraph{Evaluation metrics.}

For Evaluation Space Learning, we report mean absolute error (MAE)
and Ordering Accuracy (OA) against the reference criterion vectors.
Given a test set \(\mathcal P\), MAE is computed as
\[
\mathrm{MAE}
=
\frac{1}{|\mathcal P|K}
\sum_{P\in\mathcal P}
\sum_{k=1}^{K}
\left|\hat s_k(P)-s_k^*(P)\right|.
\]

Ordering Accuracy measures whether relative criterion-level orderings are preserved. For each criterion \(k\), a proposal pair is correct when the predicted and reference differences have the same sign:
\[
\mathrm{OA}
=
\frac{1}{|\mathcal Q|}
\sum_{(A,B,k)\in\mathcal Q}
\mathbb I\!\left[
\operatorname{sgn}\!\left(\hat s_k(A)-\hat s_k(B)\right)
=
\operatorname{sgn}\!\left(s_k^*(A)-s_k^*(B)\right)
\right],
\]
where \(\mathcal Q\) contains evaluated proposal-pair--criterion combinations. For Population Taste Distillation, since each reference preference is aggregated across multiple LLM judges, we report SP described in Section~\ref{sec:exp-individual} rather than RW-SP, as judge-specific confidence no longer provides a meaningful pair-level measure of preference strength.

\paragraph{Robustness across backbones.}
We examine whether the gains from PHD persist across backbone models by substituting the backbone across all three learning components. For Evaluation Space Learning and Population Taste Distillation, we compare the complete method with its w/o PHD counterpart under each backbone. For Individual Taste Adaptation, we compare our complete method with PP and CIPHER, the strongest baselines within their respective categories, using the same backbone. Table~\ref{tab:backbone-sensitivity} shows that PHD improves criterion estimation and population preference prediction under all four backbones. At the individual level, our complete method also outperforms PP and CIPHER for every backbone, although absolute RW-SP varies across models. These results indicate that the observed gains are not specific to the default \texttt{gpt-5.5} configuration.

\paragraph{Sensitivity to population prior.}

We isolate the role of population knowledge by fixing the \texttt{gpt-5.5} backbone and replacing only the population prior, while keeping the support feedback and individual adaptation procedure unchanged. We use the same researcher cohorts and RW-SP protocol as Section~\ref{sec:exp-individual}. We consider three existing judges as alternative priors: Auto-J~\citep{li2024generative}, a general-purpose pairwise judge; MLR-Judge~\citep{chen2025mlrbench}, which evaluates research proposals through rubric-based assessment; and SciJudge~\citep{tong2026taste}, a scientific judge for comparing scientific work. As shown in Table~\ref{tab:deployment-prior}(b), differences between population priors persist after identical individual adaptation. Sparse individual feedback therefore does not fully overwrite the population initialization, indicating that the choice of population prior continues to affect downstream personalization.

\subsection{Downstream Evaluation}
\label{sec:exp-deployment}

\paragraph{Data and protocol.}
We deploy the complete framework in an interactive proposal-exploration system to test whether learned individual taste improves downstream research-direction discovery. We evaluate 20 real researchers from quantum, computer science, statistics, and medicine, with simulated cohorts evaluated either within quantum or under cross-domain transfer, where taste learned from quantum is tested on proposals from other fields. Unlike the evaluation in Section~\ref{sec:exp-individual}, accumulated interactions dynamically shape the candidate pool in subsequent rounds.

We evaluate both real and simulated users with different comparator sets. For real users, we compare against Single Call (SC) in a matched setting with the same backbone, candidate pool, proposal-visible information, and interaction history; SC directly conditions on the history, whereas ours updates an explicit individual taste mapping.  Given the cost of real-user interaction, we restrict live comparison to this matched setting. In simulation, where additional comparisons require no further researcher interaction, we broaden baseline coverage with Profile Prompting (PP) and SynthesizeMe (SM), the strongest in-context and agentic baselines, respectively, in our simulated-user evaluation (Table~\ref{tab:individual-main}). For each comparison, we evaluate our method against one alternative method at a time using source-blind competitive-pair interleaving~\citep{radlinski2008interleaving,zhang2025airbnb}, a widely used protocol for online evaluation in ranking and recommendation systems. Proposals selected by both methods are shown once, while ranking disagreements form adjacent competitive pairs with randomized order and hidden method identity. Each proposal receives an engagement score \(e(P)\in\{0,1,2,3\}\), denoting no substantive engagement, inspection, refinement, and adoption, respectively. Further system and evaluation details are provided in Appendix~\ref{app:deployment}.

\paragraph{Evaluation metrics.}
For the \(N_{\mathrm{CP}}\) ranking disagreements, we define \textbf{Competitive-Pair Engagement Gain} as
\(
\Delta_{\mathrm{CP}}=\sum_{c=1}^{N_{\mathrm{CP}}}(e_c^{\mathrm{ours}}-e_c^{\mathrm{base}})/N_{\mathrm{CP}},
\)
measuring the average engagement advantage within directly competing proposal pairs. We also report \textbf{Win Rate}, the fraction of non-tied pairs with \(e_c^{\mathrm{ours}}>e_c^{\mathrm{base}}\). To evaluate the full ranking beyond local disagreements, we treat engagement as graded relevance and report \(\Delta_{\mathrm{NDCG}}=\mathrm{NDCG}_{\mathrm{ours}}-\mathrm{NDCG}_{\mathrm{base}}\)~\citep{jarvelin2002cumulated}.

\paragraph{Results.}
Table~\ref{tab:deployment-prior}(a) shows that learned individual taste improves proposal discovery over different baselines as the candidate pool evolves through interaction. The gains persist under cross-domain transfer, though with varying magnitude, highlighting differences in scientific taste across both individuals and research domains.

\begin{table}[t]
\centering
\caption{
\textbf{Downstream evaluation and sensitivity to population prior.}
(a) Each column compares our personalized ranker against the specified
comparator. For simulated researchers, each entry reports
\textbf{Quantum | Cross-domain}. Real-user deployment uses Single Call
(SC), while simulated cohorts additionally include Profile Prompting
(PP) and SynthesizeMe (SM).
(b) Individual RW-SP after replacing the population prior while keeping
the backbone, support feedback, and adaptation procedure fixed.
}
\label{tab:deployment-prior}

\small
\renewcommand{\arraystretch}{1.12}

\begin{minipage}[t]{0.72\linewidth}
\vspace{0pt}
\centering
\textbf{(a) Downstream evaluation}
\par\medskip

\setlength{\tabcolsep}{3pt}
\begin{tabular*}{\linewidth}
{@{\extracolsep{\fill}}lcccc@{}}
\toprule
& \multicolumn{3}{c}{Simulated researchers}
& Real researchers \\
\cmidrule(lr){2-4}
\cmidrule(lr){5-5}
Metric & SC & PP & SM & SC \\
\midrule
$\Delta_{\mathrm{CP}}\uparrow$
& $0.169 \mid 0.120$
& $0.130 \mid 0.067$
& $0.146 \mid 0.151$
& 0.125 \\
Win Rate $\uparrow$
& $0.667 \mid 0.583$
& $0.571 \mid 0.577$
& $0.641 \mid 0.635$
& 0.684 \\
$\Delta_{\mathrm{NDCG}}\uparrow$
& $0.064 \mid 0.006$
& $0.020 \mid 0.018$
& $0.037 \mid 0.044$
& 0.023 \\
\bottomrule
\end{tabular*}
\end{minipage}
\hfill
\begin{minipage}[t]{0.25\linewidth}
\vspace{0pt}
\centering
\textbf{(b) Population prior}
\par\medskip

\setlength{\tabcolsep}{3pt}
\begin{tabular}{@{}lc@{}}
\toprule
Population Prior & RW-SP $\uparrow$ \\
\midrule
Auto-J & 0.583 \\
SciJudge & 0.644 \\
MLR-Judge & 0.667 \\
\midrule
\textbf{Ours} & \textbf{0.693} \\
\bottomrule
\end{tabular}
\end{minipage}
\end{table}

\begin{table}[t]
\centering
\caption{
\textbf{Ablation studies.}
(a) Effect of PHD and privileged hindsight on the two shared components
under proposal-only inference.
(b) Effect of individual residuals and conditional rules on RW-SP.
The w/o PHD variant starts from the seed Rubric B
and a randomly initialized population prior.
}
\label{tab:ablation-studies}

\small
\renewcommand{\arraystretch}{1.12}

\begin{minipage}[t]{0.45\linewidth}
\vspace{0pt}
\centering
\textbf{(a) Privileged hindsight}
\par\medskip

\setlength{\tabcolsep}{3pt}
\begin{tabular*}{\linewidth}
{@{\extracolsep{\fill}}lccc@{}}
\toprule
Variant
& E-MAE $\downarrow$
& E-OA $\uparrow$
& P-SP $\uparrow$ \\
\midrule
w/o PHD
& 0.292 & 0.604 & 0.567 \\
w/o $\mathcal I_{\rm priv}$
& 0.242 & 0.649 & 0.533 \\
\textbf{Ours}
& \textbf{0.233} & \textbf{0.665} & \textbf{0.667} \\
\bottomrule
\end{tabular*}
\end{minipage}
\hfill
\begin{minipage}[t]{0.51\linewidth}
\vspace{0pt}
\centering
\textbf{(b) Individual adaptation}
\par\medskip

\setlength{\tabcolsep}{3pt}
\begin{tabular*}{\linewidth}
{@{\extracolsep{\fill}}lccc@{}}
\toprule
Variant
& Real $\uparrow$
& Sim.-Q $\uparrow$
& Sim.-Ctrl. $\uparrow$ \\
\midrule
Population
& 0.680 & 0.537 & 0.592 \\
+ Residual
& 0.679 & 0.573 & \textbf{0.606} \\
+ Rules
& \textbf{0.693} & \textbf{0.588} & 0.593 \\
\midrule
w/o PHD
& 0.670 & 0.540 & 0.581 \\
\bottomrule
\end{tabular*}
\end{minipage}
\end{table}

\subsection{Role of Privileged Hindsight}
\label{sec:exp-phd}

To isolate privileged hindsight, we use the shared-component evaluation from Section~\ref{sec:exp-robustness} under \texttt{gpt-5.4-mini}, comparing w/o PHD, w/o \(\mathcal I_{\rm priv}\), and Ours. The w/o PHD variant removes the distillation procedure entirely, whereas w/o \(\mathcal I_{\rm priv}\) retains the same refinement workflow but restricts the teacher to proposal-visible information. Table~\ref{tab:ablation-studies}(a) shows that proposal-visible refinement substantially improves criterion-level assessment over w/o PHD, but does not improve population preference prediction. Adding privileged hindsight yields a smaller further gain in E-MAE and E-OA but a much larger improvement in P-SP. This contrast suggests that hindsight is particularly important for learning how criterion-level differences translate into population preferences, not for criterion assessment itself.

\subsection{Understanding Individual Taste Learning}
\label{sec:exp-analysis}

\paragraph{Beyond Population Initialization.}

We first examine the two individual adaptation components under the same cohorts and RW-SP protocol as Section~\ref{sec:exp-individual}. As shown in Table~\ref{tab:ablation-studies}(b), for real researchers, the population initialization is already strong, and residual reweighting alone does not improve the aggregate result. Conditional rules yield a modest gain on real researchers and Sim.-Q, whereas residual adaptation is sufficient on Sim.-Ctrl., whose preferences are explicitly constructed from criterion-level trade-offs. This suggests that some researcher-specific preferences require context-dependent adaptation beyond fixed criterion reweighting, motivating analysis of when individual taste departs from the population prior rather than assuming uniform gains. A concrete example in Appendix~\ref{app:divergent-taste-case} shows two real users (U04 and U05) making opposing choices over the same proposal pairs despite identical frozen evaluation features, with the complete mapping recovering more of these conflicting preferences than residual-only adaptation. Removing PHD and adapting from the seed Rubric B and a randomly initialized prior further reduces RW-SP, suggesting that the PHD-derived shared structure provides a better starting point for personalization.

\begin{table}[t]
\centering
\caption{
\textbf{Replay under sequential feedback.}
Effect of replay-based rule acceptance on held-out RW-SP
as support feedback accumulates.
The ablation bypasses the cross-fitted replay gate
for accepting personal conditional rules.
}
\label{tab:replay}

\small
\setlength{\tabcolsep}{5pt}
\renewcommand{\arraystretch}{1.12}

\begin{tabular*}{0.9\linewidth}
{@{\extracolsep{\fill}}lccccccc@{}}
\toprule
Support count & 0 & 5 & 10 & 12 & 13 & 14 & 15 \\
\midrule
Ours
& 0.538 & 0.535 & 0.580 & 0.603 & 0.648 & 0.599 & 0.613 \\
w/o replay
& 0.538 & 0.535 & 0.580 & 0.575 & 0.615 & 0.578 & 0.602 \\
\bottomrule
\end{tabular*}
\end{table}

\paragraph{Replay under sequential feedback.}
We ablate the cross-fitted replay gate from Section~\ref{sec:individual-taste} under the same sequential support protocol as Section~\ref{sec:exp-individual}, while keeping the population initialization, feedback sequence, residual fitting, rule proposer, and coefficient estimation fixed. Since this analysis repeatedly evaluates adaptation over multiple support prefixes, we use gpt-5.4-mini to reduce inference cost. Table~\ref{tab:replay} shows that replay has little effect early but becomes beneficial as researcher-specific preference emerges. Held-out RW-SP remains non-monotonic, indicating that later feedback can interfere with earlier preferences and motivating replay to preserve prior evidence.

%% file: appendix.tex
\definecolor{rubricbg}{RGB}{247,248,250}
\definecolor{rubricframe}{RGB}{112,121,135}

\definecolor{learnedbg}{RGB}{246,249,247}
\definecolor{learnedframe}{RGB}{78,110,88}

\definecolor{memorybg}{RGB}{247,247,252}
\definecolor{memoryframe}{RGB}{89,91,126}

\definecolor{casebg}{RGB}{249,249,249}
\definecolor{caseframe}{RGB}{150,150,150}

\newtcolorbox{rubricbox}[1]{
    enhanced,
    breakable,
    colback=rubricbg,
    colframe=rubricframe,
    colbacktitle=rubricframe,
    coltitle=white,
    title={#1},
    fonttitle=\bfseries,
    boxrule=0.65pt,
    arc=1.5mm,
    left=3mm,
    right=3mm,
    top=2mm,
    bottom=2mm,
    before skip=7pt,
    after skip=9pt
}

\newtcolorbox{learnedbox}[1]{
    enhanced,
    breakable,
    colback=learnedbg,
    colframe=learnedframe,
    colbacktitle=learnedframe,
    coltitle=white,
    title={#1},
    fonttitle=\bfseries,
    boxrule=0.65pt,
    arc=1.5mm,
    left=3mm,
    right=3mm,
    top=2mm,
    bottom=2mm,
    before skip=7pt,
    after skip=9pt
}

\newtcolorbox{memorybox}[1]{
    enhanced,
    breakable,
    colback=memorybg,
    colframe=memoryframe,
    colbacktitle=memoryframe,
    coltitle=white,
    title={#1},
    fonttitle=\bfseries,
    boxrule=0.65pt,
    arc=1.5mm,
    left=3mm,
    right=3mm,
    top=2mm,
    bottom=2mm,
    before skip=7pt,
    after skip=9pt
}

\newtcolorbox{casebox}[1]{
    enhanced,
    breakable,
    colback=casebg,
    colframe=caseframe,
    colbacktitle=caseframe,
    coltitle=white,
    title={#1},
    fonttitle=\bfseries,
    boxrule=0.6pt,
    arc=1.5mm,
    left=3mm,
    right=3mm,
    top=2mm,
    bottom=2mm,
    before skip=7pt,
    after skip=9pt
}

\definecolor{promptbg}{RGB}{247,248,250}
\definecolor{promptframe}{RGB}{105,115,130}

\definecolor{panelbg}{RGB}{248,249,251}
\definecolor{panelframe}{RGB}{126,134,148}

\newtcolorbox{promptbox}[1]{
    enhanced,
    breakable,
    colback=promptbg,
    colframe=promptframe,
    colbacktitle=promptframe,
    coltitle=white,
    title={#1},
    fonttitle=\bfseries,
    boxrule=0.65pt,
    arc=1.5mm,
    left=3mm,
    right=3mm,
    top=2.5mm,
    bottom=2.5mm,
    before skip=7pt,
    after skip=9pt
}

\newtcolorbox{panelbox}[1]{
    enhanced,
    breakable,
    colback=panelbg,
    colframe=panelframe,
    colbacktitle=panelframe,
    coltitle=white,
    title={#1},
    fonttitle=\bfseries,
    boxrule=0.6pt,
    arc=1.5mm,
    left=3mm,
    right=3mm,
    top=2mm,
    bottom=2mm,
    before skip=7pt,
    after skip=9pt
}

\newcommand{\rubricdim}[1]{\medskip\noindent\textbf{#1}\par\vspace{2pt}}
\newcommand{\scoreanchor}[2]{\noindent\textbf{#1}\quad #2\par\vspace{2pt}}
\newcommand{\artifactfield}[1]{\noindent\textbf{#1}\quad}

\section{Privileged Hindsight Distillation Details}
\label{app:phd}

We provide additional implementation details for the four-step PHD procedure described in Section~\ref{sec:phd} and illustrated in Figure~\ref{fig:pipeline}(b). The procedure identifies informative student errors, uses privileged hindsight evidence to derive reusable corrections, and accepts artifact revisions only after leakage auditing and validation.

\subsection{Stage-Specific Failure Diagnosis}
\label{app:phd-diagnosis}

At each iteration, the analyzer compares the student rollout with the corresponding stage reference and organizes informative errors into a structured failure package. The analyzer identifies errors but does not use hindsight evidence to explain their underlying causes. Such explanations are produced by the privileged teacher in the subsequent step.

Rather than passing every disagreement to the teacher, the analyzer prioritizes failures that may yield reusable improvements. These include large prediction errors, inconsistencies between judgments and supporting evidence, explicit constraint violations, and errors recurring across multiple items. The resulting failure package is structured according to the learning task.

\paragraph{Evaluation-space assessment.} Errors are identified at the proposal and criterion levels. For each selected case, the failure package records the student's predicted score, confidence, supporting comment, evidence-gap flag, and constraint-violation flag, together with the signed error against the reference criterion score. Recurring errors across proposals are also recorded to support the identification of systematic problems.

\paragraph{Population taste prediction.} Errors are identified at the pairwise comparison level. Each selected case records the student's predicted preference, confidence, rationale, and reference preference, together with the relevant criterion coordinates and utility margin. When available, results from both comparison orientations are retained to expose orientation inconsistencies. Errors are grouped according to research type and decisive evaluation criteria to identify recurring preference failures.

The resulting package provides the teacher with structured evidence of which judgments failed and where the failures occurred. Determining why they occurred and how to correct them remains the responsibility of the privileged teacher.

\subsection{Privileged Teacher Editing}
\label{app:phd-edit}

The privileged teacher receives the current artifacts, student rollout, structured failure package, and privileged hindsight information. It examines the selected failures in light of completed methods, experimental outcomes, subsequent literature, citations, and community feedback to identify potential causes of the student's errors.

The teacher then proposes revisions to the agentic artifacts. These revisions may modify evaluation rubrics, introduce reusable memory lessons, or add and revise conditional rules. The teacher does not directly replace the student's judgments with privileged answers.

The key constraint is that hindsight may inform the discovery of a correction, but the resulting artifact must remain executable using proposal-visible information. For example, retrospective evidence may reveal that a scoring rubric systematically undervalues proposals extending an established research direction. The teacher may consequently revise the rubric to distinguish incremental extensions from substantive methodological advances. It may not encode the later experimental success of a particular proposal as an inference-time instruction.

The teacher's private diagnostic explanations are retained for analysis but are not incorporated into student-facing artifacts. Only the proposed artifact revisions are passed to the subsequent audit and validation steps.

\subsection{Artifact Leakage Audit}
\label{app:phd-leakage}

A leakage auditor examines each proposed revision before it can be accepted. The audit is applied to student-facing artifacts, including rubric instructions, memory lessons, conditional predicates, and their associated explanations. It does not constrain the privileged information used internally by the teacher to derive a revision.

The auditor rejects revisions that directly expose retrospective outcomes, citation counts, community responses, reference criterion scores, or preference labels associated with individual proposals. It also rejects references to held-out items and conditional rules that require features unavailable at the proposal stage.

For conditional rules, the auditor additionally checks that their predicates operate on proposal-visible properties and conform to the predefined predicate grammar. Instance-specific rules that effectively identify a single proposal rather than expressing a reusable judgment principle are rejected.

This distinction permits hindsight-derived knowledge to be distilled into general evaluation instructions while preventing privileged instance-level information from entering the student's inference-time artifacts. The leakage constraints and artifact output schemas remain fixed throughout the refinement process.

\subsection{Gated Validation and Artifact Acceptance}
\label{app:phd-validation}

Passing the leakage audit is necessary but insufficient for accepting a revision. A candidate artifact must also demonstrate improved performance on validation data that did not motivate the edit.

Let $\Psi_t$ denote the current artifacts and $\Psi_t'$ the candidate revision. The student is evaluated under both configurations using the same held-out validation examples. The revision is accepted only when all three conditions are satisfied.

\begin{enumerate}
\item The leakage auditor approves all modified student-facing artifacts.

\item The candidate improves the task-specific validation objective.

\item The revision does not introduce material behavioral regressions.
\end{enumerate}

For evaluation-space assessment, validation considers errors against reference criterion scores, with mean absolute error serving as the primary objective. Additional checks examine whether the revision introduces new errors in previously correct judgments.

For population taste prediction, validation uses StrictPair correctness on held-out comparisons. Regression checks examine correct-to-wrong preference flips, orientation inconsistencies, and failures concentrated in particular groups of comparisons.

The validation data exclude examples used to induce the candidate revision. A separate final evaluation set remains untouched by artifact selection and acceptance decisions.

If the candidate passes all acceptance conditions, its artifacts replace the incumbent configuration,
\[
\Psi_{t+1} =
\begin{cases}
\Psi_t', & \text{if the revision is accepted},\\
\Psi_t, & \text{otherwise}.
\end{cases}
\]

Rejected revisions are rolled back, while accepted revisions guide subsequent student rollouts. This procedure aims to retain reusable improvements that generalize beyond the errors that originally motivated the edit.

\paragraph{Evaluation-space assessment.} A revision is accepted only when the fixed leakage auditor returns a normalized risk below $0.15$, its validation MAE improves by more than $0.01$, its correct-to-wrong flip rate does not exceed $5\%$, and its net repair is positive.

\paragraph{Population taste prediction.} Candidate revisions are induced on $D_{\mathrm{induce}}$, fitted on $D_{\mathrm{rule\text{-}select}}$, and evaluated for acceptance on $D_{\mathrm{agent\text{-}val}}$. The final evaluation set $D_{\mathrm{final}}$ remains untouched.

Acceptance requires an improvement in StrictPair, a McNemar test with $p<0.05$, and a paired bootstrap $95\%$ confidence interval for the StrictPair difference that excludes zero.

\paragraph{Rule specificity.} The leakage auditor rejects rules whose triggering behavior is concentrated on a single paper or cluster, according to the configured concentration threshold.

\section{Evaluation Space Learning}
\label{app:merit-learning}

This section provides the complete criterion-annotation and proposal-time assessment instructions used to learn the shared evaluation space, together with a concrete example of how Privileged Hindsight Distillation (PHD) transforms teacher-side supervision into deployable artifacts. We first define the reference annotation rubric, then give the seed proposal-time rubric, and finally illustrate how PHD revises the rubric and student-safe memory on a held-out proposal.

\subsection{Reference Criterion Annotation}
\label{app:rubric-a}

Rubric A defines the reference annotation procedure used to construct
\[
s^*(P)=
\left(
s^*_{\mathrm{gen}},
s^*_{\mathrm{nov}},
s^*_{\mathrm{imp}},
s^*_{\mathrm{par}},
s^*_{\mathrm{dep}},
s^*_{\mathrm{comp}}
\right)
\in \{0,1,2,3\}^{6}.
\]
The six coordinates correspond to generality, novelty, importance, parsimony, depth, and algorithmic complexity. Each coordinate is assigned independently on an absolute 0--3 scale.

Unlike the deployable proposal-time assessor, Rubric A may use both proposal-visible evidence and privileged hindsight evidence. Proposal-visible evidence includes the proposal text, its explicit claims and assumptions, and literature available at proposal time. Privileged evidence includes the completed methodology and experimental findings in the corresponding paper, post-publication technical discussion, and subsequent community signals such as citations. The annotator must connect each assigned score to criterion-specific evidence rather than infer one criterion from writing quality, venue, author identity, or overall paper reputation.

\begin{rubricbox}{Rubric A: Reference Criterion Annotation}

\textbf{Input.} A research proposal, proposal-time prior literature, and teacher-side privileged hindsight evidence.

\medskip
\textbf{Evidence rule.} Use all available evidence that is relevant to the target criterion. Every assigned score must be justified by criterion-specific evidence. If multiple sources disagree, record the disagreement and select the score best supported by the overall evidence. Reduce confidence when evidence is incomplete, assumptions are not directly comparable, or adjacent score levels remain similarly plausible.

\rubricdim{Generality}
\scoreanchor{Score 0.}{The method applies only to one special setting, and its scope or applicable regime is missing or merely implied.}
\scoreanchor{Score 1.}{The method covers multiple instances or variants within one problem family, with limited variation in access, noise, or hardware assumptions.}
\scoreanchor{Score 2.}{The method covers multiple problem, input, or noise settings within one broader class, with explicit applicability assumptions.}
\scoreanchor{Score 3.}{The method covers multiple problem classes or distinct access, error, or hardware models, with explicit applicability conditions.}

\rubricdim{Novelty}
\scoreanchor{Score 0.}{The contribution is substantially the same as identified prior work, with no meaningful structural difference.}
\scoreanchor{Score 1.}{The contribution is a nearby transfer, implementation variation, or minor modification of a known method.}
\scoreanchor{Score 2.}{The contribution forms a new workflow, formulation, or structured extension by combining known mechanisms and clearly distinguishes itself from prior work.}
\scoreanchor{Score 3.}{The contribution introduces a new primitive, reduction, encoding, or formal object absent from the relevant prior-work lineage.}

\rubricdim{Importance}
\scoreanchor{Score 0.}{There is no clear evidence of field demand; the motivation is absent, generic, or primarily self-asserted.}
\scoreanchor{Score 1.}{The work addresses a useful but local subproblem, with limited evidence of broader impact.}
\scoreanchor{Score 2.}{The addressed limitation is repeatedly identified as important across multiple works, settings, or application areas.}
\scoreanchor{Score 3.}{An authoritative survey, roadmap, benchmark, or broad body of evidence identifies the problem as a central open problem or foundational bottleneck.}

\rubricdim{Parsimony}
\scoreanchor{Score 0.}{The method requires many layers, assumptions, special cases, or ad hoc components for its stated functionality.}
\scoreanchor{Score 1.}{The method is a moderately complex pipeline with several distinct construction or execution stages.}
\scoreanchor{Score 2.}{A small number of coordinated core subroutines, oracles, or assumptions achieves the stated objective.}
\scoreanchor{Score 3.}{A unified primitive, representation, or objective replaces multiple otherwise separate or ad hoc components.}

\rubricdim{Depth}
\scoreanchor{Score 0.}{The work addresses a single setting and provides little explanation of the underlying mechanism.}
\scoreanchor{Score 1.}{The work remains within one task family and provides mainly intuition or partial technical detail, with central analysis deferred.}
\scoreanchor{Score 2.}{The work unifies multiple tasks or primitives, or provides concrete analytical, experimental, or diagnostic grounding of the mechanism.}
\scoreanchor{Score 3.}{The work develops a transferable formalism supported by theorem-level, lower-bound, causal, or comparably strong mechanistic evidence.}

\rubricdim{Algorithmic Complexity}
\scoreanchor{Score 0.}{The work makes no meaningful complexity or resource claim relative to an appropriate baseline.}
\scoreanchor{Score 1.}{The work makes an informal efficiency claim without directly comparable quantitative evidence.}
\scoreanchor{Score 2.}{The work presents an explicit resource improvement relative to a domain baseline under stated assumptions and resource metrics.}
\scoreanchor{Score 3.}{The work establishes a theoretical bound or strong Pareto improvement while reasonably accounting for state preparation, oracle access, error dependence, measurement, and readout costs.}

\medskip
\textbf{Criterion-specific evidence requirements.} Generality requires evidence about covered problem classes, access models, hardware regimes, and the assumptions under which the claimed coverage holds. Novelty is assessed relative to proposal-time prior work across problem formulation, quantum primitive, reduction or encoding, complexity model, measurement or verification strategy, and assumption set. Importance requires external evidence that the target problem is recognized beyond the proposal's own motivation. Parsimony is evaluated from the complete method structure, including necessary subroutines, oracles, ancillas, measurements, post-selection steps, and special-case handling. Depth jointly measures the breadth of unification and the strength of mechanistic grounding. Algorithmic complexity compares the claimed and realized resource behavior with appropriate baselines while accounting for hidden computational costs.

\end{rubricbox}

\subsection{Seed Proposal-Time Rubric}
\label{app:rubric-b}

The deployable criterion assessor $f_{\mathrm{crit}}$ uses the same six-dimensional evaluation space and the same absolute 0--3 scale as the reference annotation, producing
\[
\hat{s}(P)=
\left(
\hat{s}_{\mathrm{gen}},
\hat{s}_{\mathrm{nov}},
\hat{s}_{\mathrm{imp}},
\hat{s}_{\mathrm{par}},
\hat{s}_{\mathrm{dep}},
\hat{s}_{\mathrm{comp}}
\right).
\]
Its information boundary, however, is different. At deployment, the assessor can access only the proposal and literature available at evaluation time. Completed outcomes, post-publication discussion, subsequent citations, and other hindsight information are unavailable.

We therefore initialize the student with the proposal-visible instruction shown below, which we refer to as \emph{seed Rubric B}. The seed rubric is an initial operationalization of the six evaluation criteria rather than a fixed definition of how the criteria must be inferred. PHD subsequently refines its score boundaries and evidence-use rules while preserving the same proposal-time information boundary.

\begin{rubricbox}{Seed Rubric B: Proposal-Time Criterion Assessment}

\textbf{Input.} A research proposal and literature available at proposal-evaluation time.

\medskip
\textbf{Evidence constraint.} Use only information available when the proposal is evaluated. Permitted evidence includes proposal claims, equations, assumptions, methodological descriptions, planned comparisons, named prior work, and proposal-time literature. Do not use completed experimental outcomes, future reception, post-publication discussion, subsequent citations, or other hindsight information.

\medskip
\textbf{Scoring rule.} Score each criterion independently on an absolute 0--3 scale. Do not infer one criterion from another, and do not reward writing style or perceived author reputation. If the available evidence is insufficient to distinguish adjacent levels, mark the evidence gap rather than silently assuming the stronger claim.

\rubricdim{Generality}
\scoreanchor{Score 0.}{The proposal addresses one special setting; its scope or applicable regimes are missing or merely suggested.}
\scoreanchor{Score 1.}{The proposal explicitly covers multiple instances within one problem family, but provides little cross-setting variation.}
\scoreanchor{Score 2.}{The proposal covers multiple problem, input, or noise settings within one class and states the relevant assumptions.}
\scoreanchor{Score 3.}{The proposal covers multiple problem classes or distinct access, error, or hardware models and states their applicability conditions.}

\rubricdim{Novelty}
\scoreanchor{Score 0.}{The proposed method is substantially the same as named prior work, with no structural difference stated.}
\scoreanchor{Score 1.}{The proposal describes a nearby transfer or minor modification of a known technique or formulation.}
\scoreanchor{Score 2.}{The proposal defines a new workflow by combining known mechanisms and states a concrete difference from named prior work.}
\scoreanchor{Score 3.}{The proposal introduces a new primitive, reduction, encoding, or formal object that is absent from the prior-work lineage described in the proposal.}

\rubricdim{Importance}
\scoreanchor{Score 0.}{The proposal provides no clear evidence of field demand; its motivation is weak, generic, or absent.}
\scoreanchor{Score 1.}{The proposal addresses a useful local subproblem but gives limited support for broader impact.}
\scoreanchor{Score 2.}{The proposal presents the limitation as repeatedly important across multiple settings, applications, or prior works.}
\scoreanchor{Score 3.}{The proposal supports the claim that the problem is a central open problem or foundational bottleneck for a broad research area.}

\rubricdim{Parsimony}
\scoreanchor{Score 0.}{The proposed method is a multi-layered pipeline with many assumptions, special cases, or ad hoc components.}
\scoreanchor{Score 1.}{The proposed method is moderately complex and contains several distinct construction or execution stages.}
\scoreanchor{Score 2.}{A small number of core subroutines or oracles achieves the stated objective with limited additional machinery.}
\scoreanchor{Score 3.}{A unified primitive, representation, or objective replaces multiple otherwise separate components.}

\rubricdim{Depth}
\scoreanchor{Score 0.}{The proposal addresses a single setting and gives little or no explanation of the underlying mechanism.}
\scoreanchor{Score 1.}{The proposal stays within one task family and provides intuition or partial detail, while its main proofs or diagnostics remain deferred.}
\scoreanchor{Score 2.}{The proposal covers multiple tasks or primitives, or gives concrete lemma, error-analysis, diagnostic, or ablation plans that ground the mechanism.}
\scoreanchor{Score 3.}{The proposal develops a transferable formalism together with explicit plans for theorem-level, lower-bound, causal, or comparably strong evidence.}

\rubricdim{Algorithmic Complexity}
\scoreanchor{Score 0.}{The proposal makes no complexity or resource claim relative to any baseline.}
\scoreanchor{Score 1.}{The proposal makes an informal efficiency claim without comparable quantitative evidence.}
\scoreanchor{Score 2.}{The proposal states an explicit improvement relative to a domain baseline, with resource metrics and assumptions.}
\scoreanchor{Score 3.}{The proposal states a theoretical bound or strong Pareto improvement and reasonably accounts for hidden state-preparation, oracle, error, measurement, and readout costs.}

\medskip
\textbf{Required structured output.} For every evaluation criterion $d$, return
\[
\hat{s}_d(P)=
\left(
\text{score},
\text{confidence},
\text{supporting comment},
\text{evidence-gap indicator},
\text{constraint-violation flag}
\right).
\]
The supporting comment must identify proposal-visible evidence for the score. The evidence-gap indicator is activated when the proposal does not provide enough information to distinguish adjacent score levels. The constraint-violation flag records a direct conflict between an assigned score and its anchor, such as assigning maximal algorithmic complexity when no resource claim is present.

\end{rubricbox}

\subsection{Learning Rubric B and Student-Safe Case Memory}
\label{app:rubric-learning}

During training, the privileged teacher observes the student's assessment $\hat{s}(P)$, the Rubric-A reference $s^*(P)$, and the supporting teacher-side evidence. Rather than transferring privileged facts into the deployable workflow, the teacher diagnoses recurring proposal-time reasoning errors and converts them into reusable student-safe supervision.

We update two complementary student-side artifacts. Systematic errors that expose ambiguity in score boundaries are translated into revisions of Rubric B. Errors that are more useful as concrete precedents are stored in a separate student-safe case memory. In both cases, hindsight-dependent facts are removed before the artifact becomes accessible to the student.

\paragraph{Rubric revisions.} Accepted rubric updates refine evidence criteria and adjacent-score boundaries. Representative learned principles include the following:
\begin{itemize}[leftmargin=1.5em]
\item several benchmark models, parameter regimes, or subcases within one task family do not automatically establish broad generality;
\item a new workflow, robustness regime, or theorem-level extension built from known primitives generally supports intermediate novelty rather than maximal novelty;
\item multiple subroutines do not necessarily imply low parsimony when the subroutines directly support one coherent mechanism;
\item explicit query, gate, depth, qubit, or error analysis can support an intermediate algorithmic-complexity score even when the strongest form of resource advantage has not been established.
\end{itemize}

\paragraph{Case memory.} Each memory entry records a proposal-visible trigger, the affected evaluation criterion, a relevant score region, and a reusable scoring lesson. Teacher-only evidence, reference outcomes, and information unavailable at proposal time are removed. At inference time, retrieved entries therefore serve as calibration precedents rather than as additional facts about the target proposal.

Candidate rubric and memory states are evaluated on the validation split using criterion-level mean absolute error,
\[
\mathcal{L}_{\mathrm{eval}}
=
\frac{1}{6|\mathcal{D}_{\mathrm{val}}|}
\sum_{P\in\mathcal{D}_{\mathrm{val}}}
\sum_{d=1}^{6}
\left|
\hat{s}_{d}(P)-s^*_{d}(P)
\right|.
\]
A candidate state is retained only when it satisfies the validation gate. Otherwise, the student is rolled back to the incumbent rubric and memory. Student-facing artifacts are additionally screened by a pattern-based leakage check before being admitted into the deployable workflow.

\begin{learnedbox}{Example Learned Rubric-B Revision}
\artifactfield{Affected criterion} Novelty

\medskip
\artifactfield{Observed failure pattern} The student repeatedly treats a new arrangement of established mechanisms as evidence of a fundamentally new primitive.

\medskip
\artifactfield{Accepted calibration rule} A new workflow, formal threshold, robustness regime, or structured extension of an existing method can justify novelty 2. Novelty 3 should be reserved for a new primitive, reduction, encoding, or formal object not already present in the relevant prior lineage.
\end{learnedbox}

\begin{memorybox}{Example Student-Safe Memory Entry}
\artifactfield{Dimension} Parsimony

\medskip
\artifactfield{Trigger} A proposal contains several proof or implementation components that all support the same central mechanism.

\medskip
\artifactfield{Calibration lesson} Parsimony is not only about having few steps; it is about whether those steps perform separate jobs or jointly support one central mechanism. A proposal with several proof or implementation components can still receive an intermediate parsimony score when the components are coordinated around the same objective.

\medskip
\artifactfield{Excluded information} Completed experimental outcomes, teacher-side hindsight evidence, the Rubric-A reference outcome, and any information unavailable at proposal time.
\end{memorybox}

\subsection{Evaluation-Space Learning Case Study}
\label{app:merit-case-study}

We illustrate the learning process on the held-out proposal \emph{Quantum Hamiltonian Descent} (arXiv 2303.\allowbreak01471). The held-out proposal is used only for analysis. Rubric revisions and memory entries are learned from analogous failure patterns in the training split rather than from this example itself. Scores are reported in the order generality, novelty, importance, parsimony, depth, and algorithmic complexity. The progress is shown in Table~\ref{tab:merit-case-study}

\begin{casebox}{Reference Annotation}
Rubric A assigns the proposal the reference criterion vector
\[
s^*(P)=(3,2,2,2,2,2).
\]

The high generality score reflects a common Hamiltonian-descent framework across smooth and non-convex optimization, digital fault-tolerant simulation, and analog Ising-machine implementations under distinct assumptions. Novelty is intermediate because the proposal introduces a distinctive workflow based on quantizing continuous-time optimization dynamics while relying on known components such as Hamiltonian simulation, product formulas, quantum Fourier transforms, and Ising embeddings. Parsimony is also intermediate because these components support one central mechanism despite a nontrivial implementation pipeline. Algorithmic complexity receives an intermediate score because the proposal states explicit resource quantities and baseline comparisons without establishing the strongest level of resource advantage with every hidden cost resolved.
\end{casebox}

\begin{casebox}{Assessment Before PHD Learning}
Using the seed Rubric B without learned case memory, the proposal-time student predicts
\[
\hat{s}_{\mathrm{seed}}(P)=(3,3,2,0,2,1).
\]

Three characteristic errors are visible. First, the student interprets the Hamiltonian formulation as a wholly new primitive and assigns novelty 3 instead of 2. Second, it counts implementation components without recognizing their shared conceptual role and assigns parsimony 0 instead of 2. Third, it treats the absence of a fully established quantum advantage as evidence of little algorithmic-complexity contribution and assigns 1 instead of 2.
\end{casebox}

\paragraph{How the teacher turns training failures into student artifacts.} On training proposals, the privileged teacher identifies analogous recurring errors. For example, when the student treats multiple prime families, decoding regimes, or circuit variants within one task as evidence of broad generality, the teacher converts the diagnosis into a proposal-visible boundary rule: multiple instances or noise regimes within one task family do not by themselves justify a high generality score. Similarly, repeated confusion between a new combination of known mechanisms and a genuinely new primitive produces the novelty calibration shown above. For parsimony, the teacher learns to distinguish raw component count from functional coherence and stores the more concrete precedent in case memory.

\begin{casebox}{Assessment After PHD Learning}
With the validation-selected Rubric B and case memory, the student predicts
\[
\hat{s}_{\mathrm{learned}}(P)=(2,2,2,1,2,1).
\]

The learned novelty boundary corrects the novelty score from 3 to 2. The parsimony prediction moves from 0 to 1, reducing but not eliminating the error. Algorithmic complexity remains underestimated. The generality score moves from the correct seed value of 3 to 2, illustrating that a transferable calibration rule can still be over-applied to an individual proposal.

The proposal-level MAE decreases from $0.667$ with the seed Rubric B to $0.500$ after learning. Across the held-out test split, absolute-score MAE decreases from 0.216 for the w/o PHD student to 0.183 for the validation-selected PHD student state. This aggregate comparison evaluates the revised rubric and memory jointly. It therefore measures improvement of the learned student state rather than isolating the causal contribution of the rubric revision or the case memory individually.
\end{casebox}

\section{Population Taste Distillation}
\label{app:PT}
\subsection{Hindsight Reference Panel}
\label{app:population-panel}

We construct population-level reference supervision using an independent multi-family LLM panel. The panel is used only to construct offline reference judgments for training and evaluation. Unlike the deployable student, panel judges may access privileged hindsight evidence in addition to proposal-time information. These privileged inputs are never exposed to the student during deployment.

\paragraph{Panel prompt.} For every proposal pair $(P_A,P_B)$, each judge independently receives the same normalized evidence bundle together with the following instruction.

\begin{promptbox}{Hindsight Reference Panel Prompt}
\small
You are an offline reference judge comparing two research proposals to construct a population-level preference reference. You receive (i) the complete text of proposals A and B, (ii) reference assessments along the six evaluation criteria, (iii) source-paper and proposal-time evidence supporting those assessments, and (iv) privileged evidence about subsequent community response, including citation function and polarity, methodological adoption, and reuse.

Using the complete evidence bundle, determine which proposal is preferable as a research direction at the population level. Consider the criterion-level assessments and community-response evidence jointly, and explain any disagreement between them. Your judgment must not depend on presentation order or proposal length. You must select exactly one winner, either A or B. Ties and uncertain outcomes are not permitted.

Return a JSON object containing the selected winner, the estimated probability that A is preferred to B at the population level, confidence in the judgment, a short rationale, the decisive factors, an explanation of how the criterion-level and community-response evidence agree or conflict, and whether a proposal-only reader could plausibly reach the same conclusion.
\end{promptbox}

\paragraph{Evidence bundle.} All judges receive the same canonical A/B orientation and the same normalized evidence bundle. The bundle combines proposal-visible information with teacher-side hindsight evidence as summarized below.

\begin{table}[t]
\centering
\small
\setlength{\tabcolsep}{5pt}
\renewcommand{\arraystretch}{1.08}
\begin{tabular}{lccc}
\toprule
Dimension & Reference & Seed & Learned \\
\midrule
Generality             & 3 & 3 & 2 \\
Novelty                & 2 & 3 & 2 \\
Importance             & 2 & 2 & 2 \\
Parsimony              & 2 & 0 & 1 \\
Depth                   & 2 & 2 & 2 \\
Algorithmic complexity & 2 & 1 & 1 \\
\midrule
MAE                     & -- & 0.667 & 0.500 \\
\bottomrule
\end{tabular}
\caption{\textbf{Evaluation-space learning on a held-out proposal.} Rubric A provides the reference coordinates. Seed and Learned show the proposal-time predictions before and after PHD learning.}
\label{tab:merit-case-study}
\end{table}

\begin{table}[t]
\centering
\small
\setlength{\tabcolsep}{5pt}
\renewcommand{\arraystretch}{1.08}
\begin{tabularx}{\linewidth}{
    >{\raggedright\arraybackslash}p{0.27\linewidth}
    >{\raggedright\arraybackslash}X
    >{\centering\arraybackslash}p{0.18\linewidth}}
\toprule
Evidence source & Content & Student-visible \\
\midrule
Proposal text
& Complete proposal claims, assumptions, equations, and methodology
& Yes \\

Proposal-time literature & Prior work and external evidence available when the proposal is evaluated & Yes \\

Criterion annotation & Six-dimensional Rubric-A assessment and its supporting evidence & No \\

Completed work & Source-paper methodology, experimental findings, and realized analyses & No \\

Community response
& Follow-up citations, scites in SciRate, citation function and polarity, adoption distance,
methodological integration, and primitive reuse
& No \\
\bottomrule
\end{tabularx}
\caption{\textbf{Evidence available to the hindsight reference panel.} Proposal-visible sources may also be accessed by the deployable workflow, whereas privileged sources remain teacher-side information.}
\label{tab:population-panel-evidence}
\end{table}

Proposal texts are truncated only when they exceed the fixed context budget. No model-specific evidence preprocessing is applied, so all panel members judge the same logical pair from the same evidence state.

\paragraph{Independent annotation.} Three model families independently annotate every proposal pair at temperature zero, namely GPT-5.5, Claude Opus 5, and Gemini 3.6 Flash. Judges cannot observe one another's outputs. Each judge returns a hard winner
\[
y_j\in\{A,B\},
\]
together with an estimated probability
\[
q_j=P_j(A\succ B),
\]
confidence, rationale, decisive factors, and an explanation of how the criterion-level and hindsight evidence support the decision.

\paragraph{Single-judge normalization.} Each response is parsed against the required output schema. The reported probability is clipped to $[0,1]$ and normalized to remain directionally consistent with the hard winner,
\[
y_j=A \Longrightarrow q_j\geq 0.51,
\qquad
y_j=B \Longrightarrow q_j\leq 0.49.
\]
A valid normalized annotation therefore cannot assign $q_j=0.5$. Malformed responses are treated as annotation failures and regenerated or excluded rather than converted into tie labels.

\paragraph{Cross-family aggregation.} The population reference label is determined only by the three hard winner labels,
\[
y_{\mathrm{panel}}(A,B)
=
\operatorname{MajorityVote}(y_1,y_2,y_3),
\qquad
y_j\in\{A,B\}.
\]
Because the panel contains three judges and every valid judge selects exactly one proposal, majority voting always produces a unique binary preference. Per-model probabilities, confidence values, rationales, and decisive factors are retained as privileged teacher-side metadata but do not alter the final reference label.

\begin{panelbox}{Reference Construction Summary}
\small
For each proposal pair, all three model families independently inspect the same hindsight evidence bundle. Each model produces one binary preference. The majority vote forms $y_{\mathrm{panel}}(A,B)$, which provides the population-level reference supervision used during Population Taste Distillation. Rich judge outputs are retained only as teacher-side diagnostic information.
\end{panelbox}

\subsection{Pairwise Instantiation of Population Taste}
\label{app:population-pairwise}

Section~\ref{sec:population-taste} defines population taste generally as a mapping from features in the shared evaluation space to research preference. In our experiments, scientific taste is observed through pairwise proposal comparison. We therefore instantiate the general mapping using the relative evaluation-space features of the two research directions.

For a proposal pair $(A,B)$, let $\hat{s}(A),\hat{s}(B)\in\mathbb{R}^{6}$ denote their estimated positions in the shared evaluation space and define the pairwise criterion difference as
\[
x_{A,B}=\hat{s}(A)-\hat{s}(B).
\]
The linear population prior uses $w_{\mathrm{pop}}^\top x_{A,B}$, where $w_{\mathrm{pop}}$ captures population-level weighting of the six evaluation criteria. PHD augments this prior with proposal-visible conditional rules, giving the pairwise instantiation of the population mapping from Section~\ref{sec:population-taste},
\[
u_{\mathrm{pop}}(x_{A,B},\mathcal{E}_{A,B})
=
w_{\mathrm{pop}}^\top x_{A,B}
+
\sum_{r\in R_{\mathrm{pop}}}
\tilde{\alpha}_r\,\phi_r(x_{A,B},\mathcal{E}_{A,B}),
\]
where $\mathcal{E}_{A,B}$ contains proposal-visible evidence for the pair, $\phi_r$ denotes the activation of rule $r$, and $\tilde{\alpha}_r$ is its reliability-adjusted contribution. The induced preference probability is $p_{\mathrm{pop}}(A\succ B)=\sigma(u_{\mathrm{pop}}(x_{A,B},\mathcal{E}_{A,B}))$.

\begin{panelbox}{From General Taste Mapping to Pairwise Supervision}
\small
The underlying object being learned is a population mapping over evaluation-space features. Pairwise comparison is the observation interface used in our task rather than the definition of population taste itself. Accordingly, the general feature representation $\hat{s}(P)$ becomes the relative representation $x_{A,B}$ when the mapping is trained and evaluated against the panel preference $y_{\mathrm{panel}}(A,B)$.
\end{panelbox}

\section{Individual Taste Adaptation Details}
\label{app:individual-details}

Individual Taste Adaptation freezes the learned evaluation-space assessor and population state and updates only the current researcher's personal state from revealed pairwise feedback. No privileged hindsight evidence, hidden researcher profile, or held-out query label is available during adaptation. The implementation first fits a population-centered residual $\Delta w_u$ by MAP logistic regression, then considers conditional personal rules only when recurring errors remain after continuous criterion reweighting. Accepted rule updates are checked by replay over the accumulated support buffer before the personal state is retained for evaluation.

\paragraph{Ablation variants.} Cold Start uses the frozen population mapping without researcher-specific updates. Residual-only fits $\Delta w_u$ while keeping the researcher-specific rule set empty. Complete uses both $\Delta w_u$ and accepted researcher-specific conditional rules. The w/o PHD variant removes PHD from the shared initialization and runs the same downstream individual-adaptation procedure starting from the seed Rubric~B and a randomly initialized population prior, while keeping the backbone, support budget, replay procedure, and adaptation mechanism unchanged.

\subsection{Replay-Based Rule Acceptance}
\label{app:individual-replay}
Candidate personal rules are evaluated over the accumulated support buffer using leave-one-out cross-fitted replay. For each support comparison $i$, we refit the individual residual $\Delta w_u$ and the candidate rule coefficient $\alpha_r$ on the remaining $n-1$ comparisons and evaluate the held-out comparison. The resulting out-of-fold predictions are used to compute three rule effects:
\emph{fixed}, the number of previously incorrect StrictPair predictions
corrected by the rule; \emph{broken}, the number of previously correct predictions made incorrect; and
\[
\mathrm{net}=\mathrm{fixed}-\mathrm{broken}.
\]
We additionally compute the change in negative log-likelihood, $\Delta_{\mathrm{NLL}} =
\mathrm{NLL}_{\mathrm{before}}-\mathrm{NLL}_{\mathrm{after}}$.

Under the default replay gate, a candidate rule is accepted only if (i) its fitted coefficient is nonzero, (ii) it activates on at least $\max(4,\operatorname{round}(k/3))$ observed comparisons, (iii) $\mathrm{net}\ge 1$, (iv) $\mathrm{broken}=0$, and (v) $\Delta_{\mathrm{NLL}}>0$. The deployed rule coefficient is obtained by aggregating the out-of-fold fitted coefficients.

The replay ablation keeps the same rule proposer, coefficient fitting, support sequence, and downstream evaluation, but bypasses conditions (ii)--(v); only the nonzero-coefficient check is retained. Thus, the ablation isolates the effect of replay-based rule acceptance rather than removing individual residual fitting or altering RW-SP. 

\subsection{A Case Study of Divergent Individual Taste}
\label{app:divergent-taste-case}

We examine two researchers, U04 and U05, who evaluate the same ten proposal pairs using identical frozen criterion features. Five pairs elicit opposing preferences, providing a direct test of whether personalized mappings can recover conflicting choices rather than reproducing a shared population prediction.

Table~\ref{tab:divergent-taste-case} shows that the cold-start mapping fails to recover both preferences on any of the five conflicting pairs. Residual-only adaptation recovers one, while the complete mapping recovers three. The improvement is particularly visible for U04, whose accuracy increases from 4/10 to 8/10. U05's accuracy, however, decreases from 7/10 at cold start to 6/10 after adaptation. This example illustrates how conditional individual rules can recover opposing preferences that residual weighting misses, while also showing that improved differentiation does not guarantee higher accuracy for every researcher.

\begin{table*}[t]
\centering
\caption{\textbf{An illustrative case of divergent scientific taste.} Two researchers evaluate the same ten proposal pairs using identical criterion features. Predictions are shown as U04/U05. Bold indicates a correct individual prediction. A star indicates that both opposing preferences are correctly recovered. Proposal titles are abbreviated for readability.}
\label{tab:divergent-taste-case}

\small
\setlength{\tabcolsep}{5pt}
\renewcommand{\arraystretch}{1.18}

\begin{tabularx}{\textwidth}{
@{}
l
>{\raggedright\arraybackslash}X
cccc
@{}
}
\toprule

Pair & Research directions (A / B) & Gold & Cold & Residual & Full \\

\midrule

\multicolumn{6}{@{}l}{
\textit{Opposing preferences between U04 and U05}
}
\\
\addlinespace[3pt]

005/008 & Eigenvalue processing / Non-unitary dynamics & B/A & A/\textbf{A} & \textbf{B}/B & \textbf{B}/B \\

009/040 & Optimal quantum simulation / Fourier-based LCHS & A/B & \textbf{A}/A & $\star$\textbf{A}/\textbf{B} & $\star$\textbf{A}/\textbf{B} \\

009/045 & Optimal quantum simulation / Circuit-efficient randomized methods & A/B & B/\textbf{B} & B/\textbf{B} & $\star$\textbf{A}/\textbf{B} \\

013/029 & Linear matrix differential equations / Gibbs expectation & B/A & A/\textbf{A} & A/\textbf{A} & $\star$\textbf{B}/\textbf{A} \\

027/034 & Quantum Hamiltonian Descent / Provably efficient algorithm & B/A & A/\textbf{A} & A/\textbf{A} & A/\textbf{A} \\

\addlinespace[3pt]
\midrule

\multicolumn{6}{@{}l}{
\textit{Aligned preferences between U04 and U05}
}
\\
\addlinespace[3pt]

013/023 & Linear matrix differential equations / Sign-embedding algorithms & B/B & A/A & A/\textbf{B} & \textbf{B}/\textbf{B} \\

028/045 & Quantum Hermite transform / Circuit-efficient randomized methods & A/A & \textbf{A}/\textbf{A} & B/B & \textbf{A}/B \\

028/049 & Quantum Hermite transform / End-to-end quantum simulation & A/A & \textbf{A}/\textbf{A} & \textbf{A}/B & \textbf{A}/B \\

029/036 & Gibbs expectation / Efficient quantum simulation & A/A & B/B & B/B & B/B \\

039/040 & BELT block encoding / Fourier-based LCHS & A/A & \textbf{A}/\textbf{A} & \textbf{A}/\textbf{A} & \textbf{A}/\textbf{A} \\

\midrule

\multicolumn{2}{@{}l}{
\textbf{Correct predictions (U04 / U05)}
}
&
--
&
4/10 \,/\, 7/10
&
4/10 \,/\, 6/10
&
\textbf{8/10} \,/\, 6/10
\\

\multicolumn{2}{@{}l}{
\textbf{Opposing-choice recovery}
}
&
--
&
0/5
&
1/5
&
\textbf{3/5}
\\

\bottomrule
\end{tabularx}

\end{table*}

\section{Data Construction and Evaluation Protocols}
\label{app:corpus-construction}

\paragraph{Source pool.} We construct the shared proposal corpus from SciRate under the search query \texttt{quantum algorithm}. After removing noise clusters and requiring clusters sufficiently rich for controlled comparison, we stratify the retained papers by topical cluster, release time, and scites. From this filtered pool, we curate two complementary sources: a broad SciRate subset for topical coverage and a classic lookup set containing historically salient quantum-algorithm papers used as a diagnostic core.

\paragraph{Proposal reconstruction.} Each retained paper is converted into a proposal-time document. We preserve the problem statement, proposed method, evaluation plan, expected contributions, and risks, while rewriting results and other post-hoc outcomes as objectives or hypotheses. Reconstruction uses full-text extraction followed by an LLM rewrite under a fixed paper-to-proposal prompt, so that downstream judges receive only information that could plausibly have been stated before the work unfolded. Controlled strengthen, weaken, and revision variants used elsewhere as comparator training signals are excluded from the Evaluation Space: absolute criterion assessment is defined only on canonical paper-derived proposals, rather than synthetic variants of the same underlying work.

\paragraph{Eligibility and paper-disjoint splits.} Both Evaluation Space Learning and Population Taste Distillation use carefully curated, paper-disjoint splits. Constructing these evaluations requires controlled proposal-time reconstruction and privileged reference construction. A retained proposal must therefore be a canonical paper-derived proposal rather than a rewrite variant, admit a single six-dimensional absolute reference under Rubric~A, support a frozen privileged-evidence package, and have its underlying paper assigned to exactly one of train, validation, or test. These constraints prevent alternative reconstructions of the same work from being treated as independent examples and ensure that privileged evidence cannot leak across splits. The validation split is used only for artifact-update gating, while the test split remains fixed throughout development.

\paragraph{Population Taste pairs.} Population Taste Distillation constructs comparisons within the same frozen proposal universe, with additional constraints on temporal comparability and informativeness. We retain temporally matched comparisons with sufficiently informative differences, while controlling proposal degree and winner imbalance when forming the broader annotation queue. The splits remain paper-disjoint across train, validation, and test: an underlying paper may participate in multiple comparisons within the same split, but never appears in another split, and no proposal pair is duplicated across splits. Privileged population references are obtained from a three-family LLM-as-a-Judge panel that observes the same proposal-visible text together with hindsight evidence. The resulting benchmark is therefore determined jointly by proposal-time reconstructability, privileged-reference availability, cross-split paper isolation, temporal comparability, and pair quality rather than by unconstrained pair-level sampling.

\paragraph{Individual taste sessions.} Individual Taste Adaptation operates under a different scarcity regime because its supervision must come directly from researchers. Offline, each of seven researchers annotates $25$ proposal pairs in one sitting: $15$ sequential support comparisons used for adaptation and $10$ held-out query pairs reserved for evaluation, each accompanied by a binary preference and confidence score. The support/query split is defined at the pair level: no query pair is used during adaptation, although the same proposal may appear in both support and query comparisons. The evaluation therefore measures generalization to unseen preference comparisons rather than to entirely unseen proposals.

The support budget is chosen to reflect the scale of supervision naturally available during interactive use. Across logged deployment sessions, a single sitting yields at most three pool rounds, corresponding to at most six reconstructed competitive preference signals, while even the most active user accumulates at most twelve such signals across sessions. The $15$-support setting therefore evaluates adaptation at approximately the same supervision scale encountered in deployment rather than assuming dense preference annotation.

The seven-researcher study is intended to test whether sparse researcher-specific feedback can improve held-out pairwise preference prediction within individuals. It is not used to estimate the population distribution of scientific taste across researchers; population-level conclusions are evaluated separately through the Population Taste benchmark above.

\paragraph{Simulated researcher cohorts.} We complement the seven real researchers with $100$ simulated researchers under the same $15$-support and $10$-query protocol. Sim.-Q contains $50$ quantum-researcher profiles derived from the publication histories of prominent researchers, while Sim.-Ctrl. contains $50$ controlled profiles whose preferences are generated by varying criterion-level trade-offs. Sim.-Ctrl. is designed to isolate preference variation that can be represented by fixed criterion reweighting, whereas Sim.-Q provides broader publication-history-derived preference patterns. Simulated cohorts are reported separately from the real-researcher results and are not treated as human evidence.

\paragraph{Researcher identifiers and privacy.} Throughout the evaluation and appendix, real researchers are represented only by pseudonymous identifiers such as U04 and U05. Researcher-specific adaptation uses observed preference feedback rather than identity-derived attributes or hidden profile information.

\section{Baseline Configurations and Metric Details}
\label{app:baseline-config}

Table~\ref{tab:baseline-config} summarizes the backbone, input representation, and feedback configuration of the evaluated baselines. Unless otherwise specified, API-based baselines use \texttt{gpt-5.5} at temperature $0$. Published methods retain their released backbones or use the adaptations specified below.

For individual taste evaluation, all methods use the same seven real-researcher sessions, with $15$ support comparisons and $10$ held-out query pairs per researcher. StrictPair requires correct predictions under both proposal orientations, and individual performance is reported using Regret-Weighted StrictPair (RW-SP).

\begin{table*}[t]
\centering
\caption{\textbf{Baseline configurations.} Backbones, input representations, and target-researcher feedback for the evaluated methods. External population judges are evaluated as alternative priors under the downstream individual-adaptation protocol. Detailed implementation settings are provided below.}
\label{tab:baseline-config}

\scriptsize
\setlength{\tabcolsep}{4pt}
\renewcommand{\arraystretch}{1.12}

\begin{tabularx}{\textwidth}{
@{}
l
l
>{\raggedright\arraybackslash}X
>{\raggedright\arraybackslash}X
c
@{}
}
\toprule
Component
& Method
& Backbone
& Input
& Support \\

\midrule

Alternative prior & Auto-J & \texttt{autoj-13b} & Proposal pair & 0 \\

Alternative prior & MLR-Judge & \texttt{claude-opus-5}, \texttt{gemini-3.6-flash} & Independent proposal scores & 0 \\

Alternative prior & SciJudge & \texttt{SciJudge-4B} & Title, abstract, date & 0 \\

\midrule

Individual & SC & \texttt{gpt-5.5} & Proposal pair, support comparisons & 15 \\

Individual & PP & \texttt{gpt-5.5} & Proposal pair, researcher profile & 15 \\

Individual & SynthesizeMe & \texttt{gpt-5.5} & Proposal pair, persona, demonstrations & 15 \\

Individual & VPL & GPT-2 encoder, released VAE & Frozen embeddings & 15 \\

Individual & FERMI & \texttt{gpt-5.5} & Personalized judge prompts & 15 \\

Individual & CIPHER & \texttt{gpt-5.5} & Proposal pair, retrieved preference notes & 15 \\

Individual & LoRe & GPT-2 encoder, rank-$8$ basis & Preference-difference embeddings & 15 \\

Individual & FSPO & \texttt{Qwen2.5-1.5B-Instruct} & Proposal pair, in-context examples & 15 \\

\bottomrule
\end{tabularx}
\end{table*}

\subsection{Alternative Population Priors}

We evaluate external scientific judges as alternative population priors for downstream adaptation. They are not used to construct the LLM-panel reference for Population Taste Distillation. Their outputs are converted into preference priors and evaluated under the same adaptation protocol.

\emph{Auto-J}~\citep{li2024generative} uses the released \texttt{GAIR/autoj-13b} model and its pairwise-tie protocol. We evaluate both proposal orientations using response-order swapping. The model receives full proposals, subject to a tokenizer maximum length of $3500$ tokens, and runs in float16 precision.

\emph{MLR-Judge}~\citep{chen2025mlrbench}
scores each proposal independently using an adapted MLR-Bench proposal rubric. The official two-model ensemble is remapped to
\texttt{claude-opus-5} and
\texttt{gemini-3.6-flash}.
We average the models' \texttt{OverallAssessment} scores for each proposal and convert the resulting pairwise difference into a soft preference prior,
\[
q_{\mathrm{ref}}
=
\operatorname{clip}_{[0,1]}
\left(
0.5+\frac{\mathrm{score}_A-\mathrm{score}_B}{18}
\right),
\]
where $\mathrm{score}_A$ and $\mathrm{score}_B$ denote the ensemble-averaged scores of proposals $A$ and $B$, respectively.

\emph{SciJudge}~\citep{tong2026taste} uses the released \texttt{OpenMOSS-Team/SciJudge-4B}. Each proposal pair is represented using its title, abstract, and date. We evaluate both $(A,B)$ and $(B,A)$ orientations. After reversing the labels from the latter orientation, we obtain two votes $v_{ab}$ and $v_{ba}$ in the original $(A,B)$ orientation, each taking a value in $\{A,B,\varnothing\}$.

We construct the soft preference prior for $A$ as
\[
q_{\mathrm{ref}}
=
\begin{cases}
0.75, & v_{ab}=v_{ba}=A,\\
0.25, & v_{ab}=v_{ba}=B,\\
0.65, & \text{only one valid vote, favoring } A,\\
0.35, & \text{only one valid vote, favoring } B,\\
0.50, & \text{otherwise}.
\end{cases}
\]
The final case includes conflicting votes and cases where both votes are missing. These values are assigned by the voting protocol rather than obtained through probability calibration.

Generation uses sampling with \texttt{do\_sample=True}, temperature $0.7$, \texttt{top\_p=0.8}, and \texttt{top\_k=20}.

\subsection{Individual-Taste Baselines}

All personalization methods receive the same $15$ support comparisons and are evaluated on the same $10$ held-out query pairs. Support memory uses arXiv abstracts, while queries use full proposal text unless otherwise specified.

The \emph{SC} baseline directly predicts pairwise preferences using the support comparisons in the input context. The \emph{PP} (Profile Prompting) baseline first summarizes the researcher using the $15$ support annotations and then uses this summary for preference prediction, rather than including all comparisons directly in the query prompt. Both variants use \texttt{gpt-5.5} at temperature $0$.

\emph{SynthesizeMe}~\citep{ryan2025synthesizeme} uses \texttt{gpt-5.5} to fit a researcher persona and select up to $4$ informative demonstrations from the $15$ text-only support preferences. We use a training-to-validation ratio of $0.7$ and evaluate $10$ search candidates.

\emph{VPL}~\citep{poddar2024vpl} trains the released variational autoencoder using frozen GPT-2 last-token embeddings. Training runs for $40$ epochs with batch size $8$, learning rate $3\times10^{-4}$, latent dimension $128$, and KL weight $10^{-3}$.

\emph{FERMI}~\citep{kim2025fermi} iteratively refines personalized judge prompts using \texttt{gpt-5.5}. We run $T=4$ optimization rounds, generating $K=3$ new prompts per round. The memory size is $4$, and the retrieval procedure selects the top $3$ entries. Optimizer sampling uses temperature $0.8$, while other API-based inference follows the default temperature of $0$.

\emph{CIPHER}~\citep{gao2024cipher} induces a short preference note from each support judgment. At query time, it retrieves the top $5$ relevant notes to condition pairwise preference prediction.

\emph{LoRe}~\citep{bose2025lore} learns a rank-$8$ shared preference basis from GPT-2 preference-difference embeddings. We use Adam with learning rate $0.05$, $2000$ joint-training steps, and $500$ few-shot adaptation steps. Researcher-specific weights $w_u$ are fitted using the $15$ support comparisons.

\emph{FSPO}~\citep{singh2025fspo} adapts \texttt{Qwen2.5-1.5B-Instruct} using LoRA and IPO on leave-one-support-out contexts. We use LoRA rank $r=8$, $2$ training epochs, learning rate $10^{-4}$, IPO coefficient $\beta=0.1$, and $4$ in-context demonstrations. Candidate preferences are scored using length-normalized continuation log-probabilities.

\subsection{Confidence Mapping for RW-SP}
\label{app:rwsp}

RW-SP uses an orientation-specific preference probability rather than the raw choice-relative confidence annotation. Let $c_i\in[0,1]$ denote the researcher's confidence in the selected direction for comparison $(A_i,B_i)$. We convert it to
\[
p_i=P(A_i\succ B_i)
=
\begin{cases}
c_i, & \text{if the researcher selects } A_i,\\
1-c_i, & \text{if the researcher selects } B_i.
\end{cases}
\]
The corresponding regret weight is
\[
d_i = 2|p_i-0.5|.
\]
Thus, $p_i=0.5$ receives zero weight, while a deterministic preference $p_i\in\{0,1\}$ receives unit weight.

When multiple persona judgments are aggregated, each choice-relative confidence is first converted to an orientation-specific probability using the same rule, and the probabilities are then averaged:
\[
q_{\mathrm{user}}=\frac{1}{m}\sum_{j=1}^{m} p_{ij}.
\]
The aggregate confidence stored by the data pipeline, $|2q_{\mathrm{user}}-1|$, is a preference-strength statistic and is not reused as $p_i$ in RW-SP.

If an orientation-specific probability is unavailable, we fall back to the hard preference label, corresponding to $p_i\in\{0,1\}$. If all comparisons have zero regret weight, RW-SP falls back to the unweighted StrictPair mean. The probabilities $p_i$ are used only for evaluation weighting. Individual residual and rule fitting use the observed binary preference labels rather than $p_i$ as soft training targets.

\section{Interactive Proposal Exploration and Deployment Evaluation}
\label{app:deployment}

\paragraph{System interface and interaction workflow.} We integrate the learned individual taste mapping into an interactive web system for exploring research directions. The system supports proposal exploration and preference feedback, allowing researchers to interact with candidate directions and refine their interests through successive interactions. The learned taste mapping is used to personalize proposal rankings based on the shared evaluation-space representation and the researcher's accumulated feedback.

\begin{figure*}[t]
\centering
\includegraphics[width=\textwidth]{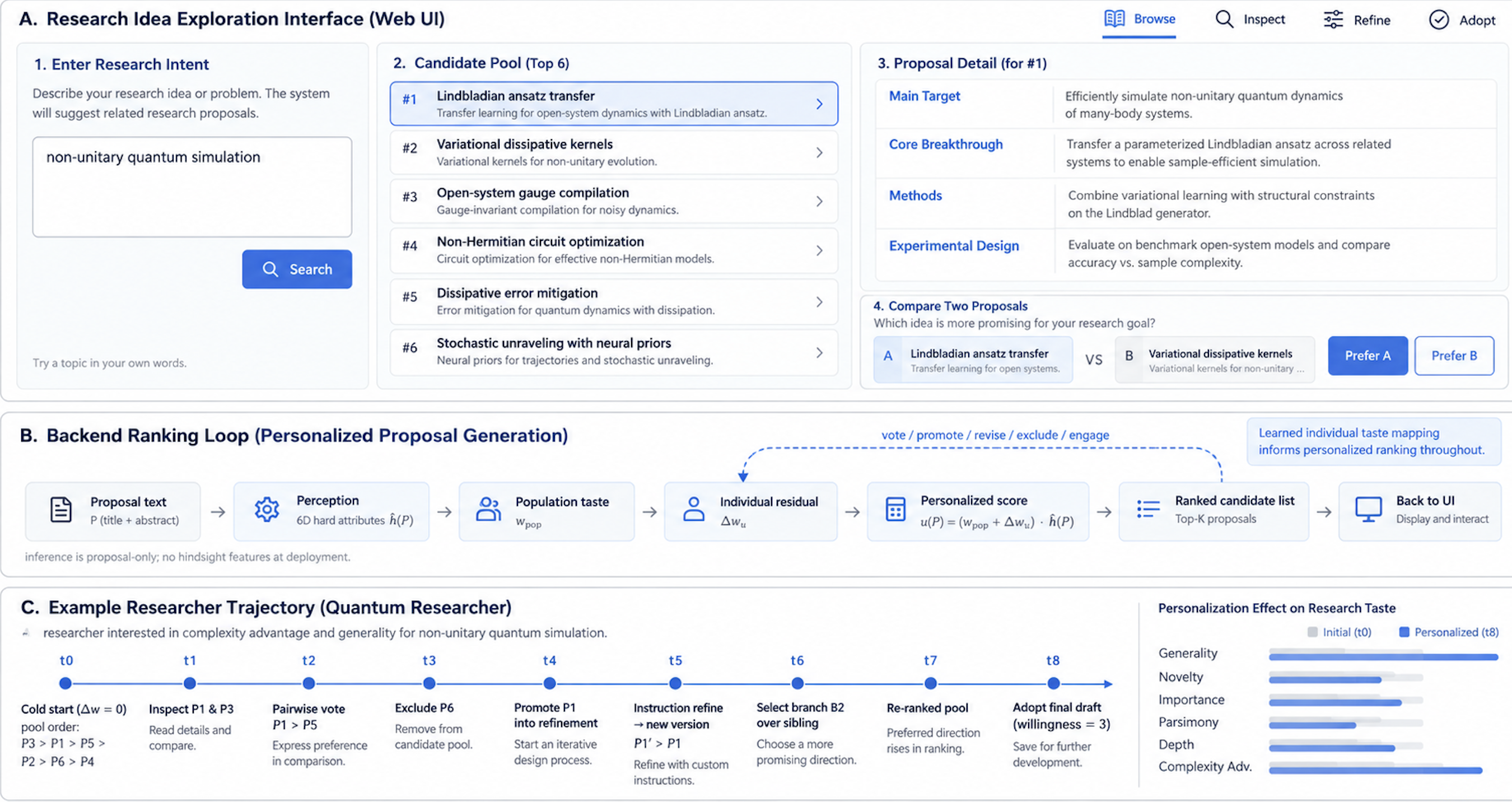}
\caption{\textbf{Interactive proposal exploration system.} The illustration presents the web interface, its main functionalities, and an example of how researchers explore candidate directions and provide feedback. The learned individual taste mapping informs personalized proposal ranking throughout the interaction process.}
\label{fig:exploration-system}
\end{figure*}

Figure~\ref{fig:exploration-system} illustrates the system interface, its main functionalities, and an example of the user interaction process. The interface provides a practical setting in which preference learning affects which research directions are presented to researchers, rather than being evaluated solely through offline pairwise predictions.

\paragraph{Researcher cohorts and evaluation protocol.} We evaluate the system with $20$ real researchers from multiple disciplines, including quantum research, computer science, statistics, and medicine. Their interaction trajectories provide behavioral evidence of which displayed proposals they choose to inspect, refine, or pursue.

To complement the real-user evaluation, we conduct offline exploration simulations with the same $100$ simulated researcher profiles described in Appendix~\ref{app:corpus-construction}: $50$ Sim.-Q profiles derived from quantum-researcher publication histories and $50$ Sim.-Ctrl. profiles generated from controlled criterion trade-offs. The simulated cohorts broaden the range of preference configurations, and we report them separately rather than treating simulated engagement as equivalent to observed human behavior.

At each interaction round, both ranking policies receive the same candidate proposal pool, proposal-visible information, and accumulated user feedback. Our personalized policy incorporates this feedback through the learned individual taste mapping, whereas the SC baseline uses the same \texttt{gpt-5.5} backbone and directly conditions on the complete interaction history in its prompt and infers researcher-specific preferences in context. Both policies use the same base model. This controlled setup holds the candidate pool, backbone, and available interaction evidence fixed, allowing the comparison to test whether explicitly learning an individual taste mapping provides benefits beyond direct in-context personalization.

\paragraph{Interaction history and supervision scale.} Across logged deployment sessions, a single sitting yields at most three pool rounds, corresponding to at most six reconstructed competitive preference signals, while even the most active user accumulates at most twelve such signals across sessions. The complete interaction history therefore remains compact enough to be included in every \texttt{SC} prompt without truncation. These trajectories also motivate the $15$-support budget used in the offline individual-taste evaluation in Appendix~\ref{app:corpus-construction}, which operates at approximately the same scale of researcher supervision.

\paragraph{Source-blind competitive-pair interleaving.} We adapt competitive-pair team-draft interleaving~\citep{radlinski2008interleaving,zhang2025airbnb} to compare the two ranking policies within a shared display.

Proposals selected by both policies are displayed only once and receive no policy-specific credit. When the policies disagree, their selections are placed together as adjacent competitive pairs. Presentation order within each pair is randomized to reduce systematic position effects. Researchers are not informed which ranking policy contributed each proposal.

This design focuses the evaluation on disagreements between the two ranking policies. Rather than comparing engagement across separately presented ranked lists, it observes user responses to competing proposals under a shared candidate pool and display.

\paragraph{Engagement signals and evaluation metrics.} We assign each displayed proposal an engagement score $e(P)\in\{0,1,2,3\}$ based on the observed interaction: $0$ indicates no substantive engagement, $1$ inspection, $2$ editing or refinement, and $3$ adoption.

Our primary metric is Competitive-Pair Engagement Gain, which measures the average engagement difference between proposals contributed by the two policies within competitive pairs,
\[
\Delta_{\mathrm{CP}}
=
\frac{1}{|\mathcal{C}|}
\sum_{c\in\mathcal{C}}
\left(
e_c^{\mathrm{ours}}
-
e_c^{\mathrm{base}}
\right),
\]
where $\mathcal{C}$ denotes the set of competitive pairs. A positive value indicates greater observed engagement with proposals contributed by the personalized policy.

We additionally report Win Rate over non-tied competitive pairs. A policy wins a pair when its contributed proposal receives a higher engagement score than the competing proposal.

To complement these pairwise comparisons, we evaluate the reconstructed rankings using NDCG, treating engagement scores as graded relevance. This metric examines whether proposals associated with greater engagement appear earlier in each policy's reconstructed ranking.

Since engagement is observed under the shared interleaved display, NDCG is interpreted descriptively. Competitive-pair outcomes remain the primary comparison between ranking policies.

Finally, engagement reflects researchers' immediate interaction with candidate directions rather than the eventual scientific outcomes of pursuing them. The deployment evaluation therefore measures usefulness for proposal exploration, not long-term scientific impact.

\section{Agentic Workflow and Prompt Specifications}
\label{app:workflow-prompts}

This appendix specifies the language-model agents and deterministic operators used to learn the three components of our framework. We follow the terminology of the main text throughout. In particular, $\hat{s}(P)$ denotes the criterion vector of research direction $P$, equivalently its position in the shared evaluation space. The six coordinates correspond to generality, novelty, importance, parsimony, depth, and algorithmic complexity.

Rubric A provides hindsight-informed reference criterion assessments, whereas Rubric B is the proposal-visible instrument available to the deployable criterion assessor. The construction of the multi-family hindsight reference panel is described separately in Appendix~\ref{app:population-panel}, and its pairwise instantiation is given in Appendix~\ref{app:population-pairwise}.

All language-model calls use temperature zero unless otherwise stated. Dynamic inputs are shown in square brackets. Retrieval, regression, orientation swapping, replay, clustering, predicate parsing, reliability estimation, validation, and acceptance decisions are deterministic operations rather than language-model agents.

\subsection{Evaluation Space Learning}
\label{app:merit-workflow-prompts}

Evaluation Space Learning estimates
\[
\hat{s}(P)=f_{\mathrm{crit}}(\mathcal I_{\mathrm{vis}}(P))
\in[0,3]^6
\]
from proposal-visible information. Its learned student-side artifacts are the deployable Rubric B and a student-safe case memory. The complete specifications of Rubric A and Rubric B are provided in Appendix~\ref{app:rubric-a} and Appendix~\ref{app:rubric-b}.

\subsubsection{Proposal-Time Criterion Assessor}

The deployable criterion assessor is the only language-model call in the forward perception path. Before each assessment, the system deterministically retrieves the $k$ nearest student-safe memory entries and places them in the prompt together with the current Rubric B.

\begin{promptbox}{Proposal-Time Criterion Assessor}
\small
\textbf{System}

You evaluate one research proposal independently on an absolute 0--3 scale along six evaluation criteria: generality, novelty, importance, parsimony, depth, and algorithmic complexity.

There is no competing proposal and no pairwise preference task. Use only proposal-visible information, the supplied evaluation rubric, and the retrieved student-safe memory. Do not use publication outcomes, citations, adoption, community response, or any information that would only become available after the proposal stage.

Return exactly one JSON object.

\medskip
\textbf{User}

Proposal identifier: [PROPOSAL ID]

[RETRIEVED STUDENT-SAFE CASE MEMORY]

[DEPLOYMENT RUBRIC B]

[PROPOSAL TEXT]

For each evaluation criterion, return:
\texttt{score} in $[0,3]$;
\texttt{confidence} in $[0,1]$;
a proposal-grounded \texttt{comment};
\texttt{evidence\_gap}; and
\texttt{constraint\_violation}.
\end{promptbox}

\subsubsection{Privileged Criterion Teacher}

A deterministic error selector compares the student's evaluation-space prediction with the Rubric-A reference and prioritizes training proposals with large criterion-level error. The selected proposal, student assessment, reference annotation, and teacher-side evidence are then passed to the privileged teacher.

\begin{promptbox}{Privileged Criterion Teacher}
\small
\textbf{System}

You are the privileged teacher for Evaluation Space Learning. You receive one research proposal, the student's six-dimensional criterion assessment, the corresponding Rubric-A reference assessment, and privileged teacher-side evidence.

Diagnose why the proposal-time student under- or over-estimated one or more evaluation criteria. Convert each diagnosis into reusable guidance for future proposal-time assessments.

Privileged evidence may motivate the diagnosis, but every emitted lesson must be executable using proposal-visible information alone. Do not copy reference scores, source-paper outcomes, evidence identifiers, citations, adoption, community-response facts, or other hindsight-dependent information into the student-facing lesson.

Teach absolute proposal-level assessment rather than pairwise preference. Return exactly one JSON object.

\medskip
\textbf{User}

[EVALUATION-CRITERION DEFINITIONS]

[PROPOSAL TEXT]

[STUDENT SCORES, CONFIDENCE, COMMENTS, AND FLAGS]

[RUBRIC-A REFERENCE ASSESSMENT --- TEACHER ONLY]

[PRIVILEGED CRITERION EVIDENCE --- TEACHER ONLY]

Return \texttt{lessons[]}. Each lesson contains:
\texttt{lesson\_type};
\texttt{evaluation\_criterion};
\texttt{proposal\_visible\_trigger};
\texttt{suggested\_check};
an optional \texttt{target\_level};
\texttt{confidence\_effect};
\texttt{leakage\_audit}; and
\texttt{student\_safe\_text}.
\end{promptbox}

\paragraph{Audit and validation.} Teacher-generated lessons pass through schema validation, leakage screening, deduplication, and conflict resolution before being attached to a candidate Rubric B or case memory. The candidate student state is then rerun on held-out proposals. An update is retained only when it improves the validation criterion; otherwise the rubric and memory are rolled back to the incumbent checkpoint. Leakage screening and predictive validation are performed separately: the former enforces the proposal-visible information boundary, whereas the latter prevents student-side updates from degrading held-out criterion assessment.

\subsection{Population Taste Distillation}
\label{app:population-workflow-prompts}

Population Taste Distillation freezes the evaluation-space assessor and learns how evaluation-space features are translated into population-level research preferences. Population supervision is provided by the hindsight reference panel described in Appendix~\ref{app:population-panel}. The deployed student can access the evaluation-space features and proposal-grounded evidence but cannot access panel labels, panel rationales, source-paper outcomes, community response, or other privileged signals.

\subsubsection{Population Taste Comparator}

The deployment comparator applies the current population mapping to a proposal pair. Conditional-rule activations are computed outside the language model and supplied as fixed inputs. Given the criterion-difference representation $x_{A,B}$, proposal-visible evidence $\mathcal{E}_{A,B}$, and the current population artifacts, the comparator instantiates the preference function defined in Section~\ref{sec:population-taste},
\[
u_{\mathrm{pop}}(x_{A,B},\mathcal{E}_{A,B})
=
w_{\mathrm{pop}}^\top x_{A,B}
+
\sum_{r\in R_{\mathrm{pop}}}
\tilde{\alpha}_r
\phi_r(x_{A,B},\mathcal{E}_{A,B}).
\]
The binary preference is determined solely by the sign of this utility:
\[
\hat y(A,B)
=
\begin{cases}
A, & u_{\mathrm{pop}}(x_{A,B},\mathcal{E}_{A,B}) > 0,\\
B, & u_{\mathrm{pop}}(x_{A,B},\mathcal{E}_{A,B}) < 0.
\end{cases}
\]
Thus, auxiliary outputs such as confidence, decisive criteria, and rationale do not override the sign-based preference decision.

\begin{promptbox}{Population Taste Comparator}
\small
\textbf{System}

You are the deployment-constrained Population Taste Comparator. Your task is to apply the learned population-level research preference policy to two research directions.

Use only the supplied evaluation-space feature vectors, proposal-grounded evidence, externally computed conditional-rule activations, and learned student-safe guidance.

Never infer or request reference-panel labels, panel rationales, source-paper outcomes, citations, adoption, community response, or any other retrospective signal. Treat all externally computed numerical quantities as fixed inputs.

Apply the supplied population mapping to obtain the scalar preference utility $u_{\mathrm{pop}}$. The preferred research direction is determined solely by the sign of this utility: positive utility selects A and negative utility selects B. Confidence, decisive criteria, and rationale are auxiliary outputs and must not alter this sign-based decision.

When the two research directions are close under the evaluation-space features, use the proposal-grounded comments, confidence values, evidence-gap flags, and constraint-violation flags only as permitted by the learned student-safe policy and conditional rules.

Return exactly one JSON object.

\medskip
\textbf{User}

Apply the learned population mapping to research directions A and B.

[CRITERION-LEVEL FEATURE VECTOR AND EVIDENCE FOR A]

[CRITERION-LEVEL FEATURE VECTOR AND EVIDENCE FOR B]

[EXTERNALLY COMPUTED CONDITIONAL-RULE ACTIVATIONS]

[LEARNED STUDENT-SAFE POPULATION POLICY AND CASE MEMORY]

Return: \texttt{preference\_utility}; \texttt{preference\_probability}; \texttt{winner} (\texttt{A} or \texttt{B}); \texttt{confidence}; \texttt{decisive\_criteria}; \texttt{used\_rule\_ids}; and a proposal-grounded \texttt{rationale}.

The returned \texttt{winner} must be consistent with
\texttt{preference\_utility}: return \texttt{A} if
\texttt{preference\_utility} $>0$ and \texttt{B} if
\texttt{preference\_utility} $<0$.
\end{promptbox}

Pair orientation swapping is used only as a deterministic robustness and validation operation. It does not participate in construction of the reference-panel label. StrictPair and related orientation-consistency statistics are likewise computed outside the language-model workflow.

\subsubsection{Privileged Population Teacher}

The privileged teacher analyzes systematic student disagreements with the population reference. It may inspect panel-derived teacher-side metadata but must convert those diagnoses into instructions that remain executable under the deployment information boundary.

\begin{promptbox}{Privileged Population Teacher}
\small
\textbf{System}

You are the privileged teacher for Population Taste Distillation. Your task is to improve the student-safe population preference policy used by a deployment-constrained comparator.

You may inspect the population reference, teacher-side panel reasoning, decisive factors, and the reported relationship between criterion-level evidence and privileged hindsight evidence. The deployment comparator cannot inspect these signals.

Translate recurring failures into reusable instructions that can be evaluated from evaluation-space features and proposal-grounded evidence alone. Never mention citations, adoption, popularity, community response, the reference panel, or the reference label in a deployed instruction.

If a reference decision depends only on retrospective information for which no proposal-visible proxy exists, abstain rather than inventing a proxy.

Propose several small and distinct policy edits. Each candidate should target one identifiable failure mode so that it can be validated independently. Return exactly one JSON object.

\medskip
\textbf{User}

Round: [ROUND INDEX]

Current population policy: [POLICY INSTRUCTIONS AND CUED MEMORY]

Student failures: [STUDENT INPUT, A/B AND B/A DECISIONS, AND PRIVILEGED REFERENCE RECORD]

Student successes that should not regress: [ANTI-REGRESSION CASES]

Return \texttt{candidates[]}. Each candidate contains:
\texttt{candidate\_id};
\texttt{diagnosis};
the complete replacement \texttt{instructions}; optional \texttt{memory\_additions}; optional \texttt{memory\_deletions}; and
\texttt{abstained\_pair\_ids}.
\end{promptbox}

Candidate packages are atomized into individual instruction or memory changes, screened for privileged-information leakage, and evaluated independently. The acceptance decision is made by a deterministic validation gate on the agent-validation split.

\subsubsection{Conditional-Rule Proposer}

A complementary teacher path proposes executable conditional rules. Privileged information may motivate a hypothesis, but every accepted rule must reduce to a predicate over one proposal's deployment-visible representation.

\begin{promptbox}{Privileged Conditional-Rule Proposer}
\small
\textbf{System}

You are a privileged hypothesis proposer for Population Taste Distillation. You do not issue the final preference judgment. Instead, you propose falsifiable conditional-taste hypotheses that a deterministic compiler and validation procedure will accept or reject.

Privileged information may inspire a hypothesis, but every deployed predicate must be evaluable independently on one research direction using its evaluation-space feature vector or a registered proposal-text extractor.

Pair-specific predicates, source-paper outcomes, citations, adoption, community response, publication venue, and reference-panel opinions are forbidden in deployed rules.

\medskip
\textbf{User}

[CURRENT FAILURE TAXONOMY]

[WORKED EXAMPLES OF PREVIOUSLY ACCEPTED RULES]

[REFINEMENT HEURISTICS]

[ERROR CLUSTER]

For each case, you receive:
\begin{itemize}
\item evaluation-space features and proposal-grounded evidence;
\item the current model prediction and margin;
\item the population-reference preference and teacher-side rationale; and
\item privileged retrospective signals.
\end{itemize}

Propose a small set of nonredundant hypotheses. Return
\texttt{hypotheses[]}, each containing:
\texttt{predicate};
\texttt{tier};
\texttt{direction} in $\{-1,+1\}$;
\texttt{failure\_mode};
a deployment-safe \texttt{explanation}; and a teacher-only \texttt{privileged\_cue}.
\end{promptbox}

Tier-1 predicates are arithmetic or Boolean functions of the six evaluation-space features and their deployment-visible evidence flags. Tier-2 predicates invoke previously registered semantic extractors over proposal-visible text. In both cases, a rule $r$ defines a single-direction activation $a_r(P)$. For a proposal pair $(A,B)$, the rule enters the comparison only through the activation difference $a_r(A)-a_r(B)$.
This preserves the general single-direction population mapping while instantiating it under the pairwise observation interface.

\subsubsection{Proposal-Text Property Evaluator}

Tier-2 predicates use a frozen proposal-text evaluator.

\begin{promptbox}{Proposal-Text Property Evaluator}
\small
\textbf{System}

Judge one research proposal against one precisely stated, present-tense property. Do not compare it with another proposal and do not speculate about future impact, adoption, reception, or other hindsight outcomes.

Return exactly one JSON object containing: \texttt{holds}; a proposal-grounded evidence span; and \texttt{confidence}.

\medskip
\textbf{User}

Property: [REGISTERED PROPERTY DESCRIPTION]

Proposal-grounded text: [PROPOSAL OR PROPOSAL-GROUNDED EVIDENCE RECORD]

If the supplied text does not clearly establish the property, return
\texttt{holds=false}.
\end{promptbox}

The property description, evaluator model, and prompt are frozen when the extractor is registered and become part of its cache key.

\subsubsection{Workflow Reflection}

The executable-rule path additionally uses a reflection agent that receives only aggregate workflow statistics. It cannot observe evaluation examples or approve its own revisions.

\begin{promptbox}{Population-Rule Workflow Reflection}
\small
\textbf{System}

You improve the conditional-rule generation workflow using aggregate round-level statistics.

You may revise the proposer preamble, worked examples, failure taxonomy, refinement heuristics, or case-selection strategy. You may not revise the predicate grammar, information-boundary contract, validation criterion, or output schema.

You never observe the evaluation examples themselves.

\medskip
\textbf{User}

[NUMBER OF PROPOSED AND TYPE-CHECKED HYPOTHESES]

[VALIDATION SURVIVAL RATE AND HELD-OUT GAIN]

[CLUSTER COVERAGE AND REDUNDANCY]

[COMMON REJECTION REASONS]

[CURRENT PROPOSER ARTIFACTS]

Diagnose the largest current weakness and propose one bounded revision. Return:
\texttt{diagnosis};
\texttt{change\_kind};
the corresponding replacement or addition; and
\texttt{change\_note}.
\end{promptbox}

The incumbent and revised proposer configurations are subsequently evaluated under identical budgets by deterministic operators. The reflection agent cannot approve its own revision.

\subsection{Individual Taste Adaptation}
\label{app:individual-workflow-prompts}

The numerical adaptation and replay gate are described in Appendix~\ref{app:individual-details}. This section records the language-model prompts used only to propose conditional personal rules and proposal-text properties after residual fitting leaves recurring support errors.

\subsubsection{Evaluation-Space Personal Rule Proposer}

The first agentic rule path searches for conditional preferences expressible within the learned evaluation space.

\begin{promptbox}{Evaluation-Space Personal Rule Proposer}
\small
\textbf{System}

You diagnose why a personalized scientific-taste model still mispredicts a specific researcher's choices after a six-dimensional weight residual has already been fitted.

Continuous criterion reweighting has therefore already been attempted. Do not propose that an evaluation criterion should simply receive more or less global weight. Instead, propose a conditional rule that activates only under a repeated pattern shared by the remaining errors.

Each proposed trigger must be evaluable independently on one research direction from its evaluation-space feature vector. State only the preferred direction of the effect; rule strength will be estimated statistically outside the language model.

A rule that activates almost everywhere is a disguised global weight update. A rule that activates only on one example is an anecdote. Return only predicates under the frozen executable grammar.

\medskip
\textbf{User}

Researcher: [USER ID]

Fitted individual weighting: [POPULATION WEIGHTS PLUS PERSONAL RESIDUAL]

Cross-fitted support errors that remain after residual fitting: [PER-SIDE EVALUATION-SPACE FEATURES, USER CHOICE, MODEL CHOICE, AND MARGIN]

Shared trigger pattern: [PATTERN]

Propose at most [BUDGET] conditional rules.
\end{promptbox}

This proposer is invoked only after sufficient individual support has accumulated and a repeated residual-error pattern can be identified.

\subsubsection{Proposal-Text Personal Property Proposer}

If no evaluation-space conditional rule survives replay, the second agentic path may propose a proposal-visible property outside the representational span of the six predefined evaluation criteria.

\begin{promptbox}{Proposal-Text Personal Property Proposer}
\small
\textbf{System}

A personalized criterion-weight residual and evaluation-space conditional rules have already failed to explain the remaining choices. Propose a reusable property of one research proposal's text that captures the unresolved taste pattern.

The property must describe a dimension of research preference rather than a research topic, proposal identity, or pair identifier. It must describe what the proposal presently states, not its future impact, adoption, reception, or other hindsight outcome.

\medskip
\textbf{User}

Researcher: [USER ID]

Fitted individual weighting: [WEIGHTS]

Remaining support errors: [EVALUATION-SPACE FEATURES, PROPOSAL-GROUNDED TEXT, USER CHOICES, AND OPTIONAL USER-PROVIDED RATIONALES]

Propose at most [BUDGET] properties. Each property must:
\begin{itemize}
\item be answerable for one research direction independently;
\item plausibly generalize across research topics;
\item separate a nontrivial subset of the support interactions rather than activating universally or on one proposal only; and
\item avoid citations, adoption, reception, hidden user information, and other hindsight-dependent signals.
\end{itemize}

Return \texttt{extractors[]}, each containing:
\texttt{name};
\texttt{description};
\texttt{direction};
\texttt{explanation}; and
\texttt{why\_these\_cases}.
\end{promptbox}

\subsubsection{Property Rewrite Agent}

If a proposed proposal-text property fails the statistical gate, a rewrite agent may inspect only the rejection diagnostics and the same observed support history. No additional labels are revealed.

\begin{promptbox}{Personal Property Rewrite Agent}
\small
\textbf{System}

A previously proposed personal-property extractor failed statistical validation. Propose a substantively different taste property rather than renaming the failed property.

\medskip
\textbf{User}

[REMAINING CROSS-FITTED ERRORS]

[FAILED PROPERTY DESCRIPTIONS]

[ACTIVATION COUNTS, FIXED/BROKEN COUNTS, NLL CHANGE, AND REJECTION REASONS]

Propose new present-tense, per-proposal properties that separate the unresolved cases and activate on enough of the complete support buffer to estimate their effect.

Topic labels, pair memorization, hindsight outcomes, and hidden researcher information are forbidden. Return the same extractor schema as the initial proposal-text property proposer.
\end{promptbox}

\subsection{Deterministic Components}
\label{app:deterministic-components}

For clarity, the following operations contain no language-model prompt: nearest-neighbor case retrieval, error ranking, logistic fitting of $w_{\mathrm{pop}}$ and $\Delta w_u$, proposal-order swapping, StrictPair computation, predicate parsing, rule-activation computation, coefficient fitting, reliability updates, cross-fitted replay, policy atomization, leakage screening, validation-gate decisions, rollback, and held-out evaluation.

This separation ensures that language-model agents propose or diagnose candidate textual artifacts, whereas acceptance and numerical parameter estimation remain externally controlled.